%% file: main.tex
\documentclass[conference]{IEEEtran}
\IEEEoverridecommandlockouts

\usepackage{cite}
\usepackage{amsmath,amssymb,amsfonts}
\usepackage{algorithmic}
\usepackage{graphicx}
\usepackage{textcomp}
\usepackage{xcolor}
\usepackage{amsmath} 
\usepackage{amsthm}
\usepackage{booktabs} 
\usepackage{array}
\usepackage{graphicx} 
\usepackage{multirow}
\usepackage{color}
\usepackage{booktabs}        
\usepackage{adjustbox}      
\usepackage{siunitx}        
\usepackage{multirow}       
\usepackage{array}  
\usepackage{makecell} 
\usepackage{float}
\usepackage{subcaption}
\usepackage{hyperref} 
\usepackage{enumitem}
\usepackage{colortbl}
\usepackage{placeins}
\usepackage{comment}
\usepackage{pifont}
\usepackage{url} 
\usepackage{lscape}
\usepackage{array} 
\usepackage[linesnumbered,vlined,ruled,noend]{algorithm2e}
\usepackage{algorithm2e}
\usepackage{placeins}
\usepackage{tcolorbox}
\usepackage{xcolor}
\usepackage{tabularx}

\newtcolorbox{graybox}{
  colback=gray!10,    
  colframe=gray!50,   
  boxrule=1pt,        
  arc=0pt,            
  boxsep=5pt,         
  left=6pt, right=6pt 
}

\definecolor{revisionblue}{named}{black}
\newcommand{\revision}[1]{\textcolor{revisionblue}{#1}}

\def\BibTeX{{\rm B\kern-.05em{\sc i\kern-.025em b}\kern-.08em
    T\kern-.1667em\lower.7ex\hbox{E}\kern-.125emX}}

\definecolor{bluelabel}{HTML}{6A8DBE}
\definecolor{greenlabel}{HTML}{9AB557}

\makeatletter
\newcommand{\linebreakand}{%
  \end{@IEEEauthorhalign}
  \hfill\mbox{}\par
  \mbox{}\hfill\begin{@IEEEauthorhalign}
}
\makeatother

\newcommand{\paperTheTable}{}
\newcommand{\paperTheFigure}{}
\newcommand{\paperTheHTable}{}
\newcommand{\paperTheHFigure}{}

\makeatletter
\AtBeginDocument{%
  \let\paperTheTable\thetable
  \let\paperTheFigure\thefigure
  \@ifundefined{theHtable}{}{\let\paperTheHTable\theHtable}%
  \@ifundefined{theHfigure}{}{\let\paperTheHFigure\theHfigure}%
}
\makeatother

\begin{document}

\newcounter{mainTable}
\newcounter{mainFigure}
\setcounter{mainTable}{\value{table}}
\setcounter{mainFigure}{\value{figure}}

\setcounter{table}{0}
\setcounter{figure}{0}
\renewcommand{\thetable}{R\paperTheTable}
\renewcommand{\thefigure}{R\paperTheFigure}

\makeatletter
\@ifundefined{theHtable}{}{%
  \renewcommand{\theHtable}{R\paperTheHTable}%
}
\@ifundefined{theHfigure}{}{%
  \renewcommand{\theHfigure}{R\paperTheHFigure}%
}
\makeatother

\setcounter{table}{\value{mainTable}}
\setcounter{figure}{\value{mainFigure}}
\let\thetable\paperTheTable
\let\thefigure\paperTheFigure
\makeatletter
\@ifundefined{theHtable}{}{\let\theHtable\paperTheHTable}
\@ifundefined{theHfigure}{}{\let\theHfigure\paperTheHFigure}
\makeatother


\input{contents/main_text_arxiv}

\clearpage

\input{contents/appendix}

\end{document}

%% file: contents/main_text_arxiv.tex
\title{\textsc{Text2SQL-Flow}: A Robust SQL-Aware Data Augmentation Framework for Text-to-SQL}

\author{
\IEEEauthorblockN{
Qifeng Cai$^{\dagger}$,
Hao Liang$^{\dagger}$,
Chang Xu$^{\dagger}$,
Tao Xie,
Wentao Zhang$^*$,
Bin Cui$^*$
}
\IEEEauthorblockA{
\textit{Peking University, Beijing, China}\\
}
\IEEEauthorblockA{
qfcai@stu.ecnu.edu.cn, \{hao.liang, cxu25\}@stu.pku.edu.cn,\\
\{taoxie, wentao.zhang, bin.cui\}@pku.edu.cn
}
}

\maketitle

\renewcommand{\thefootnote}{\fnsymbol{footnote}}
\footnotetext[2]{Equal contribution.\enspace\enspace$^{*}$Corresponding authors.}
\renewcommand{\thefootnote}{\arabic{footnote}}

\begin{abstract}
The data-centric paradigm has emerged as a pivotal direction in artificial intelligence (AI), relying on high-quality training data. This shift is especially critical in the Text-to-SQL task, where the scarcity, limited diversity, and structural simplicity of existing datasets constrain model performance. To address these challenges, we propose \textsc{Text2SQL-Flow}, a SQL-aware data augmentation framework that systematically generates large-scale, semantically valid, and structurally diverse Text-to-SQL pairs from limited seed data. 
Our framework operates along six augmentation dimensions and integrates an end-to-end pipeline featuring auxiliary database selection, SQL executability verification, natural language (NL) question generation, NL–SQL correspondence verification, and chain-of-thought (CoT) reasoning trace generation.
Leveraging this framework, we construct \textsc{SQLFlow}, a high-quality dataset comprising 75,386 annotated examples. 
We demonstrate the utility of \textsc{SQLFlow} in both fine-tuning and prompt-based settings:  
(1) For open-source large language models (LLMs), fine-tuning with \textsc{SQLFlow} enhances the problem-solving capabilities. Under the same data budget, models trained on \textsc{SQLFlow} achieve competitive performance gains across multiple benchmarks.  
(2) For closed-source LLMs, we propose a masked alignment retrieval method that leverages \textsc{SQLFlow} as both a knowledge base and the training data for the retrieval model. This approach enables structure-aware example matching by modeling fine-grained alignments between NL questions and SQL queries.  Experimental results show that our retrieval strategy outperforms existing example retrieval methods, highlighting the dual importance of \textsc{SQLFlow}’s high-quality data and our novel retrieval technique. 
Our work establishes a scalable, data-centric foundation for advancing Text-to-SQL systems and underscores the indispensable role of structured, high-fidelity data in modern AI development.
Our code is available at \url{https://github.com/TechNomad-ds/Text2SQL-Flow}.
\end{abstract}


\section{Introduction}
\label{intro}

In recent years, the data-centric artificial intelligence (AI) paradigm has garnered increasing attention~\cite{zha2025data, jakubik2024data}. Unlike traditional algorithm-centric approaches that primarily focus on advancing model architectures and algorithms, many AI applications are increasingly constrained by the availability of high-quality data. Vast amounts of data remain underutilized, despite their immense potential value. From a data-centric perspective, high-quality, diverse, and well-structured training data plays a central role in advancing AI~\cite{haukkala2022data}.

\begin{table*}[t]
    \centering
    \caption{Comparison of SQL query complexity between the original seed datasets and the datasets augmented by \textsc{Text2SQL-Flow}. Rows prefixed with ``SQLFlow-'' show augmented dataset stats; the bottom row shows overall \textsc{SQLFlow} statistics.}
    \label{tab:overall_statistics}
    \resizebox{\textwidth}{!}{
    \begin{tabular}{l|rccccccc}
       \toprule
    \multirow{2}{*}{\textbf{Dataset}} & 
    \multirow{2}{*}{\textbf{Size}} & 
    \multicolumn{4}{c}{\textbf{Query Feature Presence per SQL \%}} & 
    \multicolumn{3}{c}{\textbf{Feature Count per SQL}} \\
    \cmidrule(lr){3-6} \cmidrule(lr){7-9}
    & & \textbf{Window Func.} & \textbf{Set Op.} & \textbf{Subquery} & \textbf{GROUP BY} & \textbf{\# CASE} & \textbf{\# WHERE} & \textbf{\# JOIN} \\
       \midrule
       Spider-train & 8,659 & 0.00 & 6.07 & 15.59 & 22.60 & 0.00 & 0.86 & 0.71 \\
       SQLFlow-Spider & 30,502 & 14.37 & 9.49 & 40.06 & 37.67 & 0.22 & 1.84 & 0.96 \\
       \midrule
       BIRD-train & 9,428 & 0.04 & 0.23 & 6.71 & 9.16 & 0.09 & 1.25 & 0.90 \\
       SQLFlow-BIRD & 33,163 & 12.39 & 7.06 & 36.28 & 29.86 & 0.26 & 2.36 & 1.10 \\
       \midrule
       EHRSQL-train & 18,472 & 11.99 & 0.23 & 67.72 & 20.72 & 0.02 & 4.13 & 0.42 \\
       SQLFlow-EHRSQL & 11,721 & 21.65 & 7.47 & 66.86 & 34.14 & 0.16 & 4.32 & 0.95 \\
       \midrule
       \textbf{\textsc{SQLFlow}} & 75,386 & 14.63 & 8.11 & 42.57 & 33.68 & 0.23 & 2.46 & 1.02 \\
    \bottomrule
    \end{tabular}
    }
\end{table*}

The Text-to-SQL task is currently facing an urgent demand for high-quality data. Although the structured query language (SQL) enables powerful and precise database interactions, its complex syntax and steep learning curve limit its accessibility for non-expert users. 
The Text-to-SQL task aims to translate natural language (NL) questions into executable structured query language (SQL) queries, allowing users without specialized knowledge to access databases conveniently~\cite{kanburouglu2024text}. This capability shows broad application prospects in intelligent analysis, business intelligence platforms, and enterprise data services. With the rise of LLMs, significant progress has been made in Text-to-SQL research. 

\begin{figure}[h] 
\centering
\includegraphics[width=0.8\columnwidth]{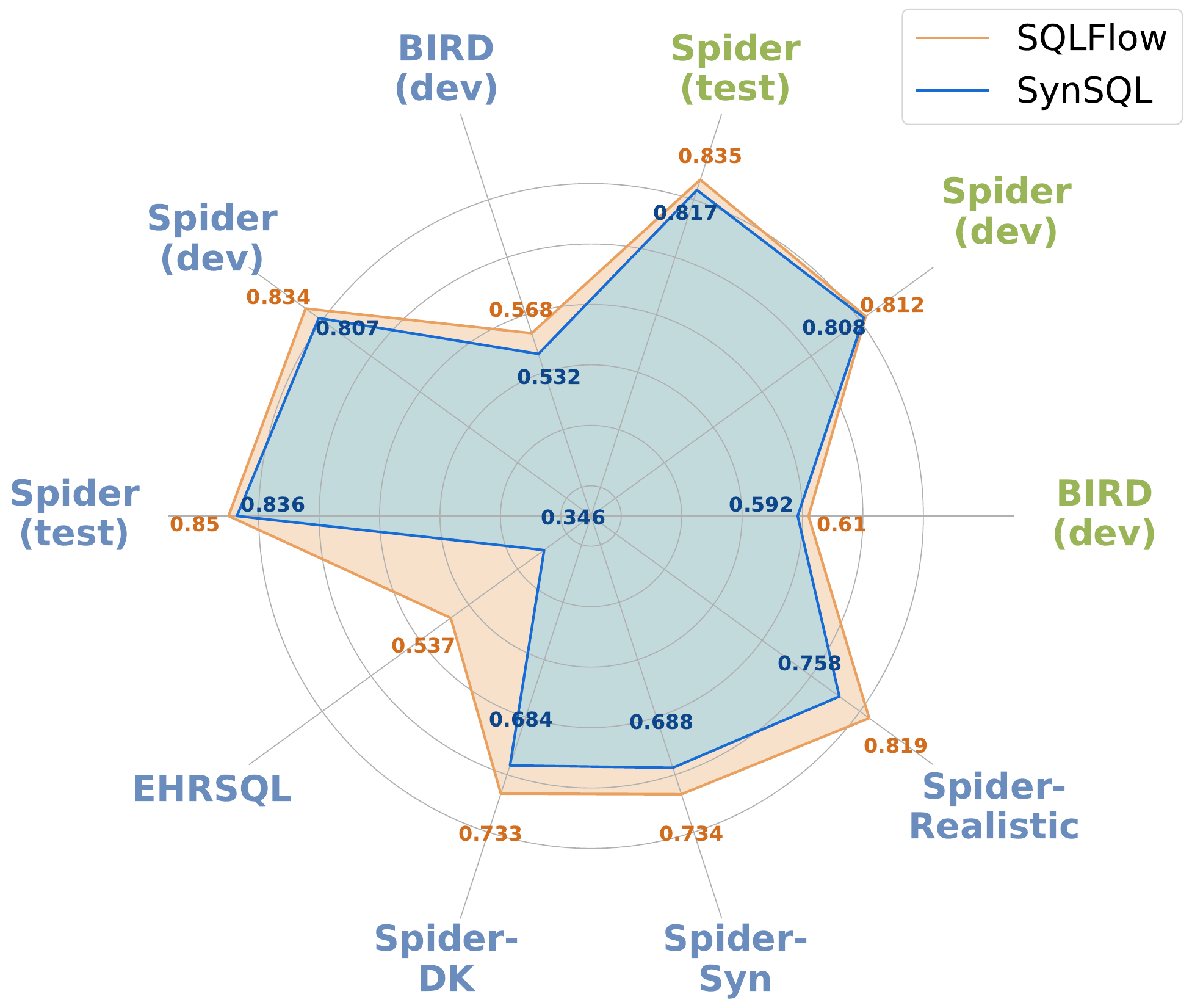}
\caption{Performance comparison of models trained on SQLFlow vs. SynSQL~\cite{li2025omnisql} (both at 75,386 samples). \textcolor{bluelabel}{Blue labels} indicate results for fine-tuned open-source models, while \textcolor{greenlabel}{green labels} indicate results of closed-source model with few-shot retrieval for prompt construction.}
\label{fig:radar_performance_comparision}
\end{figure}

Existing methods are mainly divided into two categories: \textit{(1) fine-tuning-based methods}~\cite{gao2023text, gao2024xiyan, li2024codes, sheng2025slm} and \textit{(2) prompt-based methods}~\cite{wei2022chain, tai2023exploring, li2023resdsql, pourreza2024din}. 
In both categories of methods, high-quality data plays a pivotal role. Fine-tuning-based methods rely on data pairs to enable LLMs to learn the mapping from NL questions to SQL queries, thereby enhancing the model's reasoning capability. In prompt-based methods, where model weights are inaccessible, in-context learning becomes key to improving performance, and the quality of retrieved few-shot examples is critical. Data can serve as reference examples in a knowledge base or be used to train retrieval models for obtaining more representative few-shot samples~\cite{liu2023comprehensive, nan2023enhancing, gao2023text}.

Since the Text-to-SQL task relies on NL questions and corresponding SQL queries, efficiently acquiring large-scale, high-quality data remains a fundamental challenge. Manual collection not only struggles to guarantee quality but is also limited in scale. Existing public datasets (e.g., Spider~\cite{yu2018spider}, BIRD~\cite{li2024can}) mainly rely on manual annotation, which leads to limited size and narrow data coverage. Recent studies~\cite{yang2024synthesizing, li2025omnisql} have explored the potential of automatically synthesizing Text-to-SQL datasets at scale using LLMs. \revision{However, these approaches often overlook the rich structural and semantic information already embedded in existing curated Text-to-SQL datasets. These datasets, which are typically carefully curated and execution-verified, provide a strong foundation for systematic and scalable data expansion. Failing to fully leverage such high-quality resources may result in suboptimal data efficiency and unnecessary redundancy in data generation.}

\label{intro_diversity}
\revision{
Data diversity in Text-to-SQL can be characterized along two dimensions.  \textit{Distribution-level diversity} refers to variations in underlying database schemas, data domains, and result distributions, and is typically introduced by incorporating databases from different sources or domains. In contrast, \textit{query-level diversity} captures the structural and compositional variations of SQL queries, such as aggregation patterns, join complexity, nested subqueries, and logical constraints.
}

\revision{
To address these challenges, we propose \textsc{Text2SQL-Flow}, a robust and scalable Text-to-SQL data augmentation framework that systematically expands existing seed data derived from original databases and their corresponding SQL queries. To enhance distribution-level diversity, our framework first introduces auxiliary databases by selecting databases with varying structural complexities and data domains, thereby enriching the overall data landscape.
Building upon seed SQL queries from both original and auxiliary databases, \textsc{Text2SQL-Flow} defines six data augmentation dimensions spanning data value transformations, query structure modifications, and complexity enhancements. Leveraging LLMs, the framework synthesizes diverse and semantically valid augmented SQL queries. A series of specialized operators is then employed to perform SQL executability filtering, NL question generation, NL–SQL alignment filtering, prompt construction, and chain-of-thought (CoT) reasoning path generation. These operators collaboratively ensure the quality, correctness, and diversity of the generated data.
Through the combination of LLM-driven generation and automated filtering mechanisms, \textsc{Text2SQL-Flow} is capable of producing large-scale, high-quality, and diverse Text-to-SQL datasets from limited original data, while maintaining strong scalability and extensibility across heterogeneous databases. Furthermore, a unified database interaction layer is adopted to abstract database-specific details, enabling efficient SQL execution and schema access across multiple relational database systems.
}

Based on our SQL augmentation pipeline, a high-quality Text-to-SQL dataset comprising 75,386 entries called \textsc{SQLFlow} is constructed. Using \textsc{SQLFlow}, we explore the applications of data in Text-to-SQL tasks. For open-source LLMs, the augmentation framework can provide higher-quality training data. The included CoT can further assist models in improving their problem-solving capabilities. After fine-tuning open-source models with data generated by our framework, we achieved excellent performance on multiple benchmarks, validating the framework's effectiveness. As shown in Figure~\ref{fig:radar_performance_comparision}, under the same data scale, the model trained by \textsc{SQLFlow} achieves higher performance than SynSQL~\cite{li2025omnisql} across all benchmarks, attaining 56.8\% on the BIRD-dev dataset. For closed-source models, we adopt prompt-based methods, retrieving few-shot examples to provide references for the LLM in solving problems, thereby achieving broad applicability across various scenarios. Existing systems typically select examples from limited public datasets, which have significant limitations in semantic diversity and structural coverage. Therefore, the knowledge base for retrieval can also be constructed using the data generated by our framework. However, at the retrieval method level, current mainstream retrieval strategies often focus merely on surface-level semantic similarity, neglecting the deep syntactic and logical relationships between NL questions and SQL queries. To address this, we propose an improved masked alignment retrieval method for example selection. This method learns the masked alignment relationships between NL questions and SQL queries during the training phase and achieves fine-grained structure-aware matching during the inference phase, enhancing the relevance and effectiveness of retrieved examples. The training data for this retrieval model also utilizes the data constructed by our framework. 
As shown in Figure~\ref{fig:radar_performance_comparision} shows that our retrieval method achieves competitive performance on benchmarks, with 61.0\% on the BIRD-dev and 83.5\% on the Spider-dev, outperforming other methods.

\begin{figure*}[t] 
\centering
\includegraphics[width=\textwidth]{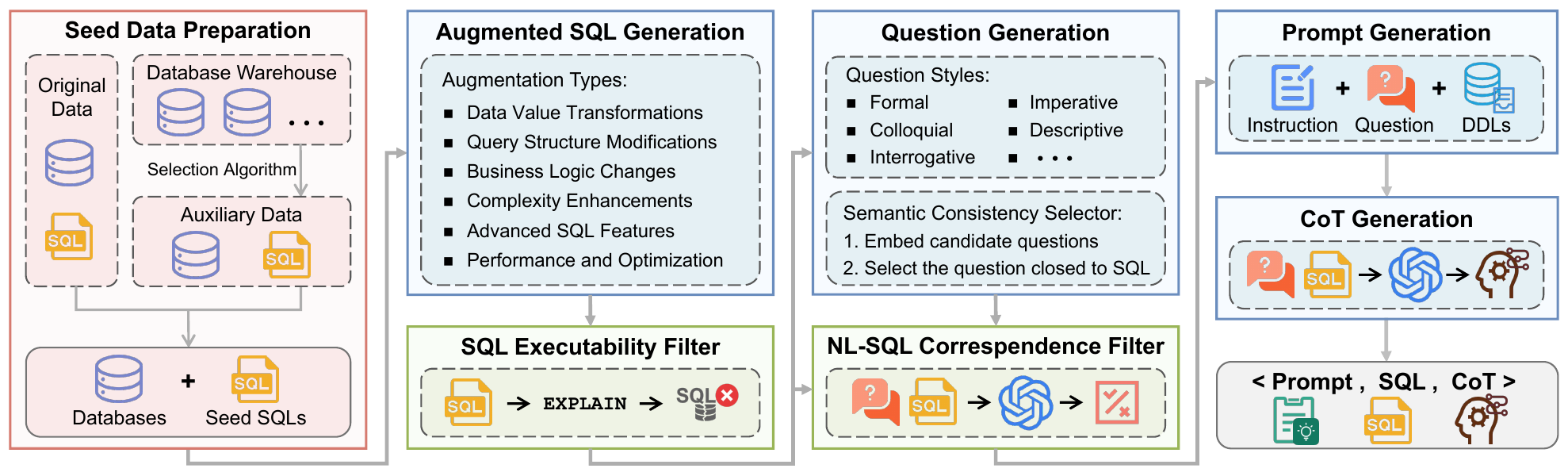}
\caption{Overall framework of our work. 
}
\label{fig:overall_framework}
\end{figure*}

Our contributions are summarized as follows:
\begin{itemize}

    \item \revision{We propose \textsc{Text2SQL-Flow}, a robust Text-to-SQL data augmentation framework that uses existing Text-to-SQL pairs as seeds to systematically generate semantically valid and structurally diverse augmented samples along six dimensions efficiently. An end-to-end automated pipeline is designed, integrating modules for auxiliary database selection, SQL augmentation generation, SQL executability filter, NL question generation, NL-SQL alignmentation filter, prompt generation, and CoT generation, ensuring high-quality generated data. }

    \item Based on \textsc{Text2SQL-Flow}, we generate a high-quality Text-to-SQL dataset named \textsc{SQLFlow}, comprising 75,386 samples. The data is efficiently generated, exhibiting high diversity and quality. Experiments on open-source LLMs show that fine-tuning with \textsc{SQLFlow} enhances model performance, achieving competitive results efficiently on benchmark evaluations. Notably, it attained 61.0\% on the BIRD-dev dataset and 85.1\% on the Spider-test dataset, outperforming other models trained with datasets of similar scale.
    
    \item  For scenarios involving closed-source models, we utilize \textsc{SQLFlow} as a few-shot example knowledge base and further propose a masked alignment retrieval method. By learning fine-grained alignment relationships between NL questions and SQL queries, our method enables more accurate example matching. Experiments show that our retrieval model, trained using \textsc{SQLFlow}, enhances the Text-to-SQL accuracy of closed-source models, outperforming existing retrieval methods based on semantic similarity, achieving 61.0\% on the BIRD-dev dataset and 83.5\% on the Spider-dev dataset.
\end{itemize}

\section{Related Works}

\subsection{Text-to-SQL Techniques}

Early rule-based approaches map NL questions to SQL using handcrafted templates~\cite{li2014constructing,mahmud2015rule,zelle1996learning}, but their reliance on rigid rules severely limits generalization across domains and schemas.
Neural sequence-to-sequence models~\cite{zhong2017seq2sql,guo2019towards,wang2019rat,chen2021shadowgnn} introduce encoder–decoder architectures with attention and schema-aware representations to model the alignment between NL and database structures. However, these methods still struggle with robustness when generalizing to unseen or heterogeneous schemas.
Pre-trained language models such as BERT~\cite{devlin2018bert,yin2020tabert} and T5~\cite{raffel2020exploring} further improve Text-to-SQL by enhancing semantic alignment between NL utterances and database schemas~\cite{fu2023miga,gu2023few}. Nonetheless, their performance heavily depends on large amounts of annotated data and degrades in low-resource or cross-domain settings.

Recently, the emergence of LLMs has opened new directions for Text-to-SQL. Through sophisticated prompting strategies, LLMs can generate accurate SQL queries even in zero-shot or few-shot settings~\cite{dong2023c3,liu2023comprehensive}. Techniques including prompt engineering~\cite{gao2023text}, task decomposition~\cite{pourreza2024din}, and self-correction have further enhanced their reliability and generalization capabilities, demonstrating considerable promise for cross-domain and real-world applications.

\subsection{Data-Centric AI for Text-to-SQL}
Data-centric approaches for Text-to-SQL focus on leveraging existing data and synthetic data generation. 
For existing data, prior work improves schema understanding via schema linking and structural representations~\cite{nahid2025rethinking, cao2024rsl, li2023resdsql, li2024multisql}, along with preprocessing and cleaning techniques~\cite{gao2024xiyan, pourreza2025reasoning, glass2025extractive, kothyari2023crush4sql} that refine schema information and inter-table relationships.

Synthetic data generation methods are generally divided into SQL-to-question and question-to-SQL paradigms. SQL-to-question approaches~\cite{guo2018question, wang2021learning, wu2021data, zhong2020grounded} generate natural language questions from SQL queries using templates or rules, producing high-fidelity pairs but suffering from limited scalability and difficulty in handling complex queries. Question-to-SQL methods~\cite{yang2021hierarchical, wei2022chain, yu2020grappa, yu2018syntaxsqlnet} generate questions first and then infer SQL using Text-to-SQL models; while more natural, these methods are prone to error propagation.
Recent work seeks to mitigate these issues. Sense~\cite{yang2024synthesizing} adopts prompt chaining to jointly generate schema, question, and SQL with semantic consistency, but relies on costly closed-source models. OmniSQL~\cite{li2025omnisql} instead decouples the generation pipeline and leverages open-source or lightweight LLMs with CoT reasoning, offering a more scalable and reproducible solution.


\section{Methodology}
\label{sec:method}

\revision{Despite the growing interest in the Text-to-SQL task, its development is fundamentally constrained by the limited availability of high-quality SQL data.
For a given Text-to-SQL system, the user typically provides an existing database $d_o$, which serves as both the available database resource and the target database on which the Text-to-SQL task is performed.
Although there exists a set of manually annotated SQL queries $\mathcal{Q}_o$ over $d_o$, its scale is usually limited.
The objective is therefore to improve the model’s performance on the Text-to-SQL task over $d_o$.
However, due to the high cost of expert annotation, $\mathcal{Q}_o$ is small in scale, making it insufficient to rely solely on it for learning robust and generalizable query patterns.
Nevertheless, $\mathcal{Q}_o$ is typically of high quality, with accurate semantics and strong alignment to the underlying database schema, making it a reliable foundation for data expansion.
As a result, how to effectively leverage $\mathcal{Q}_o$ to construct large-scale and diverse SQL data remains challenging.}

\label{intro_start_diversity}
\revision{When expanding the scale of SQL data, diversity should not be overlooked. 
SQL diversity can be characterized along two complementary dimensions. 
\textit{Distribution-level diversity} captures variations in database schemas, data domains, and underlying data distributions. 
Such diversity enhances domain coverage and mitigates overfitting to database-specific patterns. 
Meanwhile, \textit{query-level diversity} reflects structural and compositional variations of SQL queries. 
Both dimensions are crucial for constructing high-quality augmented SQL datasets.
Motivated by these insights, we propose a joint data augmentation framework, \textsc{Text2SQL-Flow} (shown in Figure~\ref{fig:overall_framework}), which simultaneously enhances distribution-level diversity and query-level diversity under data-scarce settings.
Based on this framework, we further design a data generation pipeline to construct large-scale, high-quality, and diverse SQL data.}

\subsection{Distribution-level Diversity Augmentation}
\label{sec:auxiliary_database_selection}

\revision{Distribution-level diversity captures variations in database schemas, data domains, and underlying data distributions.
When the original database $d_o$ is fixed, an effective way to improve distribution-level diversity is to introduce additional databases as auxiliary resources.
Each auxiliary database is typically accompanied by its own set of existing SQL queries, which naturally expands the diversity of SQL patterns.
Moreover, newly introduced databases bring novel data domains and distinct data distributions, thereby enriching the space of feasible SQL augmentations.
This is because SQL queries are inherently shaped by the underlying data distributions they operate on.
However, the number of available databases is often large, making it impractical to incorporate all of them.
We therefore construct a database warehouse $\mathcal{D}_w$ consisting of publicly available and accessible databases.
Each database $d \in \mathcal{D}_w$ is associated with a corresponding SQL set $\mathcal{Q}(d)$.
Given resource constraints, it is infeasible to utilize the entire warehouse.
Instead, we select a subset of databases from $\mathcal{D}_w$ as auxiliary databases, denoted as $\mathcal{D}_a$, to enhance distribution-level diversity.
Our key principle for selecting $\mathcal{D}_a$ is that, under a limited budget, the chosen auxiliary databases should cover a wide range of schema complexities.}

\revision{For each candidate database $d \in \mathcal{D}_w$, we compute a set of complexity metrics that depend solely on the database schema structure, including the number of tables $T(d)$, the total number of columns $C(d)$, and the foreign key density
$\text{FK\_density}(d) = \frac{\text{FK}(d)}{\max(T(d), \epsilon)}$
where $\epsilon$ is a small smoothing constant used to ensure numerical stability when the number of tables is small.
To avoid inconsistencies in scale across different metrics and the need for manually specified weights, we compute the percentile rank of each complexity metric over the database warehouse $\mathcal{D}_w$.
Specifically, for any metric $x(d)$, its percentile rank is defined as
$r_x(d) = \frac{1}{|\mathcal{D}_w|} \sum_{d' \in \mathcal{D}_w} \mathbb{I}\!\left[x(d') \le x(d)\right]$
where $\mathbb{I}[\cdot]$ denotes the indicator function.
Accordingly, we obtain $r_T(d)$, $r_C(d)$, and $r_{FK}(d)$.
Based on these rankings, we define the schema complexity score of a database as
\begin{equation}
    \text{score}(d) = r_T(d) + r_C(d) + r_{FK}(d).
\end{equation}
We then sort all candidate databases according to $\text{score}(d)$ and select $k$ quantile points at uniform intervals over the score distribution, thereby determining $k$ auxiliary databases.}

\revision{We then construct a combined corpus consisting of the SQL queries from the original database $\mathcal{Q}_o$ and those from the auxiliary databases $\bigcup_{d \in \mathcal{D}_a} \mathcal{Q}(d)$.
Although this process substantially increases distribution-level diversity, the SQL queries themselves remain constrained by the structural patterns inherent in the original datasets.
Therefore, an augmentation strategy over existing SQL queries is required to further expand query-level diversity.}

\subsection{Query-level Diversity Augmentation}

\subsubsection{\textbf{SQL Augmentation}}
\label{sql_augmentation}

Existing approaches to SQL data synthesis often focus on generating queries from scratch, such as via templates or random construction, overlooking the untapped potential of high-quality SQL queries already available. This limits the effective utilization of existing resources. In contrast, such high-quality queries can be repurposed through data augmentation to generate diverse yet semantically equivalent or closely related augmented SQL data. Moreover, SQL queries are inherently tied to their underlying database schemas. Constraints such as table or column names and foreign key relationships are implicitly encoded in valid SQL queries. 
Augmenting from existing queries naturally preserves schema consistency, ensuring that the generated augmented data remains syntactically valid and semantically aligned with the original data distribution. This yields augmented data that better reflects real-world query patterns while maintaining fidelity to the database schema.

A key challenge in data augmentation lies in producing meaningful and diverse augmented data from existing SQL queries. Generating schema-consistent and semantically coherent queries not only enriches the training data but also enhances the model’s robustness and adaptability to a wide range of query formulations.
To this end, we propose a data augmentation framework that leverages LLMs to generate high-quality augmented SQL data. Thanks to extensive exposure to SQL snippets during the pre-training process, LLMs have acquired strong priors over SQL syntax, semantics, and common query patterns, making them well-suited for generating syntactically valid and semantically faithful SQL data.
However, a critical limitation remains: existing public SQL datasets often exhibit limited query patterns and relatively low structural complexity. When used as prompts or seeds, such simplistic queries can constrain the diversity and sophistication of the generated variants, hindering coverage of real-world usage scenarios. To overcome this, we design a set of targeted augmentation strategies that preserve the core semantics of the original queries while systematically increasing their structural complexity and stylistic variation. This enables the construction of a more comprehensive and representative Text-to-SQL dataset. Six augmentation strategies are proposed to generate SQL queries in different directions, which can be summarized as follows:
\begin{itemize}
    \item \textbf{Data Value Transformations (DVT)}: Modify query parameters to refine result precision, including filtering conditions, date ranges, numeric thresholds, sorting criteria, limit values, and aggregation boundaries (e.g., \texttt{GROUP BY} at different temporal granularities).
    \item \textbf{Query Structure Modifications (QSM)}: Alter the structural form of queries by rewriting aggregation queries as window functions (and vice versa), introducing subqueries or common table expressions (CTEs), substituting \texttt{JOIN} with \texttt{EXISTS} or \texttt{IN}, and switching between correlated and uncorrelated subqueries to improve flexibility and maintainability.
    \item \textbf{Business Logic Changes (BLC)}: Adapt queries to alternative analytical objectives or business contexts, such as shifting across domains (e.g., sales to inventory), adjusting data granularity (e.g., daily to monthly), changing analytical perspectives (e.g., profit to cost), or modifying evaluation metrics (e.g., sum to average).
    \item \textbf{Complexity Enhancements (CE)}: Increase query complexity by introducing additional filtering conditions, joining extra tables, incorporating \texttt{CASE} expressions for conditional logic, or embedding data validation and quality checks.
    \item \textbf{Advanced SQL Features (ASF)}: Employ advanced SQL constructs, including partitioned window functions, set operations (\texttt{UNION}, \texttt{INTERSECT}, \texttt{EXCEPT}), recursive CTEs, and pivot/unpivot operations, to extend the technical expressiveness of queries.
    \item \textbf{Performance and Optimization (PO)}: Optimize query execution by applying optimization hints, restructuring queries to leverage indexes, adopting efficient query patterns, and refining \texttt{WHERE} clauses to balance performance and resource utilization.
\end{itemize}

\label{value_sample}
To ensure the validity of the generated augmented SQL queries, we incorporate randomly sampled database values into the prompts to improve semantic grounding. Each prompt consists of the following components: \textit{(1) Instruction} (\(I_{\text{sql}}\)): a directive guiding the LLM to generate SQL queries that satisfy realistic analytical requirements;
\textit{(2) Database Schema} (\(S\)): the full set of \texttt{CREATE TABLE} statements describing all relations in the database;  
\revision{\textit{(3) Database Values} (\(V\)): Values are randomly sampled from multiple rows in each table to construct meaningful query predicates; any sampled row containing null values is replaced.}
\textit{(4) Original SQL query} (\(s_i^{\text{orig}}\)): the query to be augmented; and \textit{(5) Augmentation Direction} (\(c\)): one of six predefined augmentation strategies. 
Let \(C = \{c_1, \dots, c_6\}\) denote the set of augmentation strategies. For each original query \(s_i^{\text{orig}}\), we uniformly sample a strategy \(c \sim \text{Uniform}(C)\), draw representative database values \(V\) from the relevant tables, and construct an augmentation prompt \(P_{\text{aug}}(I, S, s_i^{\text{orig}}, V, c)\). The augmented query is then produced by the LLM as:
\begin{equation}
s_i^{\text{aug}} = \text{LLM}\big(P_{\text{aug}}(I_\text{sql}, S, V, c, s_i^{\text{orig}})\big)
\end{equation}
where \(P_{\text{aug}}(\cdot)\) denotes the prompt construction process, and \(\text{LLM}(\cdot)\) represents the stochastic query generation by LLM.

\begin{figure*}[t] 
\centering
\includegraphics[width=0.85\textwidth]{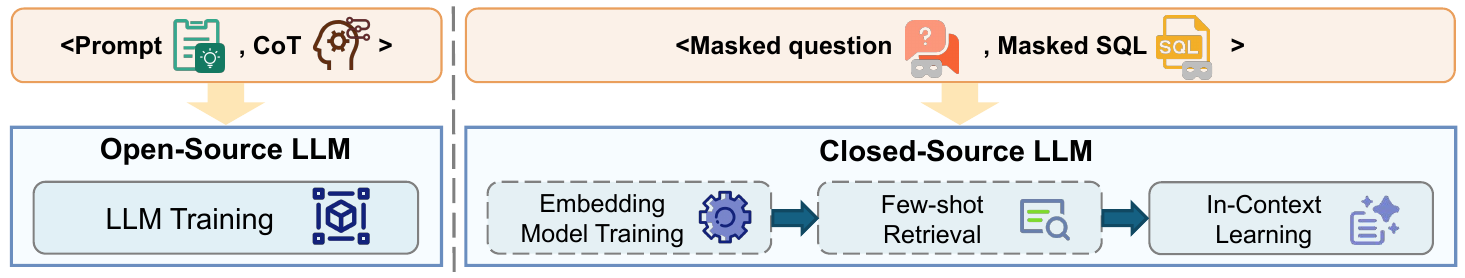}
\caption{Augmented data utilization for open-source and closed-source LLMs}
\label{fig:data_utilization}
\end{figure*}

\subsubsection{\textbf{SQL Executability Filter}}
\label{executability_filter}
\revision{Not all generated SQL queries are executable, as some may contain syntax errors or reference non-existent database objects. 
Therefore, we apply an SQL executability filter to remove invalid candidates.
Under a fixed database schema and a read-only setting without user-defined functions, we verify executability at the plan level. 
Specifically, for each generated SQL query, we check whether the database can successfully parse the query and generate an execution plan using \texttt{EXPLAIN} (e.g., \texttt{EXPLAIN QUERY PLAN} in SQLite).
A query is retained if plan generation succeeds; otherwise, it is discarded.
Since this verification does not execute the query, it is insensitive to runtime variability such as cold caches or transient system load.
This design choice avoids unintentionally biasing the augmented SQL set toward ``fast'' queries, while ensuring that all retained queries are logically executable under the fixed schema, which is sufficient for our SQL data augmentation task.}

\subsubsection{\textbf{Question Generation}}
\label{question_generation}

After obtaining the valid augmented SQL queries, the next step is to generate semantically equivalent NL questions. To ensure linguistic diversity in the synthetic data, it is essential to incorporate a wide range of stylistic instructions. Inspired by OmniSQL~\cite{li2025omnisql}, we analyze real-world user queries along multiple linguistic dimensions and identify eleven common stylistic categories, grouped into four high-level aspects:

\begin{itemize}
    \item \textbf{Tone and Formality (TF)}: levels of formality, including formal and colloquial styles
    \item \textbf{Syntactic Structure and Intent (SSI)}: sentence types such as imperative, interrogative, and declarative
    \item \textbf{Information Density and Clarity (IDC)}: degrees of conciseness, descriptiveness, vagueness, and metaphorical expression
    \item \textbf{Interaction Patterns (IP)}: interaction modes such as role-playing and procedural interactions
\end{itemize}

The first two categories capture cases where user intent is unambiguous but expressed through varying tones or syntactic forms. In contrast, vague and metaphorical styles involve ambiguous or figurative language, often requiring external or contextual knowledge for accurate interpretation.

To synthesize NL questions, we design structured prompts for LLMs comprising the following four components:  
\textit{(1) Instruction} (\(I_{\text{nl}}\)): a directive instructing the LLM to translate a given SQL query into an NL question;  
\textit{(2) Augmented SQL Query} (\(s_i^{\text{aug}}\)): the SQL query to be translated;  
\textit{(3) Database Schema} (\(S\)): the complete schema provided as \texttt{CREATE TABLE} statements; and  
\textit{(4) Target Language Style} (\(\tau\)): a stylistic variant randomly sampled from a predefined set \(\mathcal{T} = \{\tau_1, \dots, \tau_{11}\}\), where each style is defined with a description and illustrative examples.

For each SQL query \(s_i\), we sample a style \(\tau \sim \text{Uniform}(\mathcal{T})\) and assemble the prompt as \(P_{\text{nl}}(I, s_i^{\text{aug}}, S, \tau)\). The LLM then generates an NL question:
\begin{equation}
q_i = \text{LLM}\big(P_{\text{nl}}(I_\text{nl}, S, \tau, s_i^{\text{aug}})\big).
\end{equation}

To ensure both semantic fidelity and linguistic diversity, we generate \(k\) candidate questions \(\{q_{i1}, \dots, q_{ik}\}\) for each \(s_i^{\text{aug}}\) by independently sampling a new style \(\tau\) for each generation. 
\label{selector}
\revision{Among these candidates, we select the candidate question that maximizes semantic consistency with the underlying SQL query, while maintaining agreement across stylistic variants.} The final synthetic dataset is constructed as \(\mathcal{D} = \{(q_i, s_i^{\text{aug}})\}_{i=1}^{N}\), where \(N\) is the number of distinct SQL queries. This approach yields a dataset with rich linguistic variation and broad semantic coverage, enhancing both robustness and generalization in downstream applications.

\subsubsection{\textbf{NL--SQL Correspondence Filter}}
\label{alignment_filter}
\revision{Even when an SQL query is executable, it may not faithfully reflect the intent of its paired NL question.
To ensure semantic consistency between the NL question and the corresponding SQL query, we introduce a NL--SQL correspondence filter.
Specifically, given a database schema $S$, a natural language question $q_i$, and its augmented SQL query $s_i^{\text{aug}}$, we employ LLM to assess whether the two express the same query intent:
\begin{equation}
\mathbb{I}_{\text{align}}(S, s_i^{\text{aug}}, q_i)
=
\mathbb{I}_{\text{LLM}}(S, s_i^{\text{aug}}, q_i),
\end{equation}
where $\mathbb{I}_{\text{LLM}}(\cdot) \in \{0,1\}$ indicates whether semantic equivalence is confirmed.
Only question--SQL pairs with $\mathbb{I}_{\text{align}} = 1$ are retained for subsequent use.}

\subsubsection{\textbf{Prompt Generation}}
For open-source LLMs, problem-solving capabilities can be enhanced through supervised fine-tuning (SFT), a process in which the model learns to map prompt inputs \( p_i \) to the corresponding outputs. To facilitate reliable Text-to-SQL generation, a well-structured prompt ($p_i$) incorporates not only the NL question ($q_i$) but also the database schema ($S$) and a specific task instruction ($I_\text{prop}$) that guides the model. The process of prompt construction is formally defined as follows:
\begin{equation}
    p_i = P_{\text{prop}}(I_\text{prop}, S, q_i)
\end{equation}
where $P_{\text{prop}}(\cdot)$ denotes the function that synthesizes the final prompt from its constituent components.

\subsubsection{\textbf{Chain-of-Thought Generation}}
To train high-quality LLMs capable of solving complex problems, supervising models using only the final SQL outputs is often insufficient. 
Chain-of-thought (CoT) traces generated by stronger models decompose the question-solving process into a sequence of smaller, manageable subproblems, enabling step-by-step reasoning. 
Such CoT supervision provides informative learning signals and can be leveraged to improve the overall reasoning performance of weaker LLMs.
In the context of Text-to-SQL translation, CoT traces typically involve the following steps: 
(1) analyzing the intent of the NL question, 
(2) interpreting the database schema, 
(3) identifying relevant tables and columns, 
(4) filtering necessary information, 
(5) selecting appropriate SQL operators, and 
(6) incrementally constructing the SQL.

To generate CoT reasoning traces, we employ an LLM with strong reasoning capabilities. The prompt design includes 
\textit{(1) an instruction} ($I_\text{cot}$), 
\textit{(2) the database schema} ($S$), 
\textit{(3) the generated NL question} ($q_i$), and 
\textit{(4) the augmented SQL query} ($s_i^{\text{aug}}$). 
Guided by this prompt, the LLM produces a complete reasoning chain that encompasses intermediate reasoning steps as well as the final SQL query, which can be expressed as:
\begin{equation}
cot_i = \text{LLM}\big(P_{\text{cot}}(I_\text{cot}, S, s_i^{\text{aug}}, q_i )\big)
\end{equation}
where $P_{\text{cot}}(\cdot)$ denotes the CoT prompt construction process. 
During validation, the generated SQL query is extracted from the reasoning chain. 
A CoT trace is accepted as a valid solution only if the execution result of the generated SQL matches that of the reference SQL on the given database.

\begin{figure}[t] 
\centering
\includegraphics[width=0.95\columnwidth]{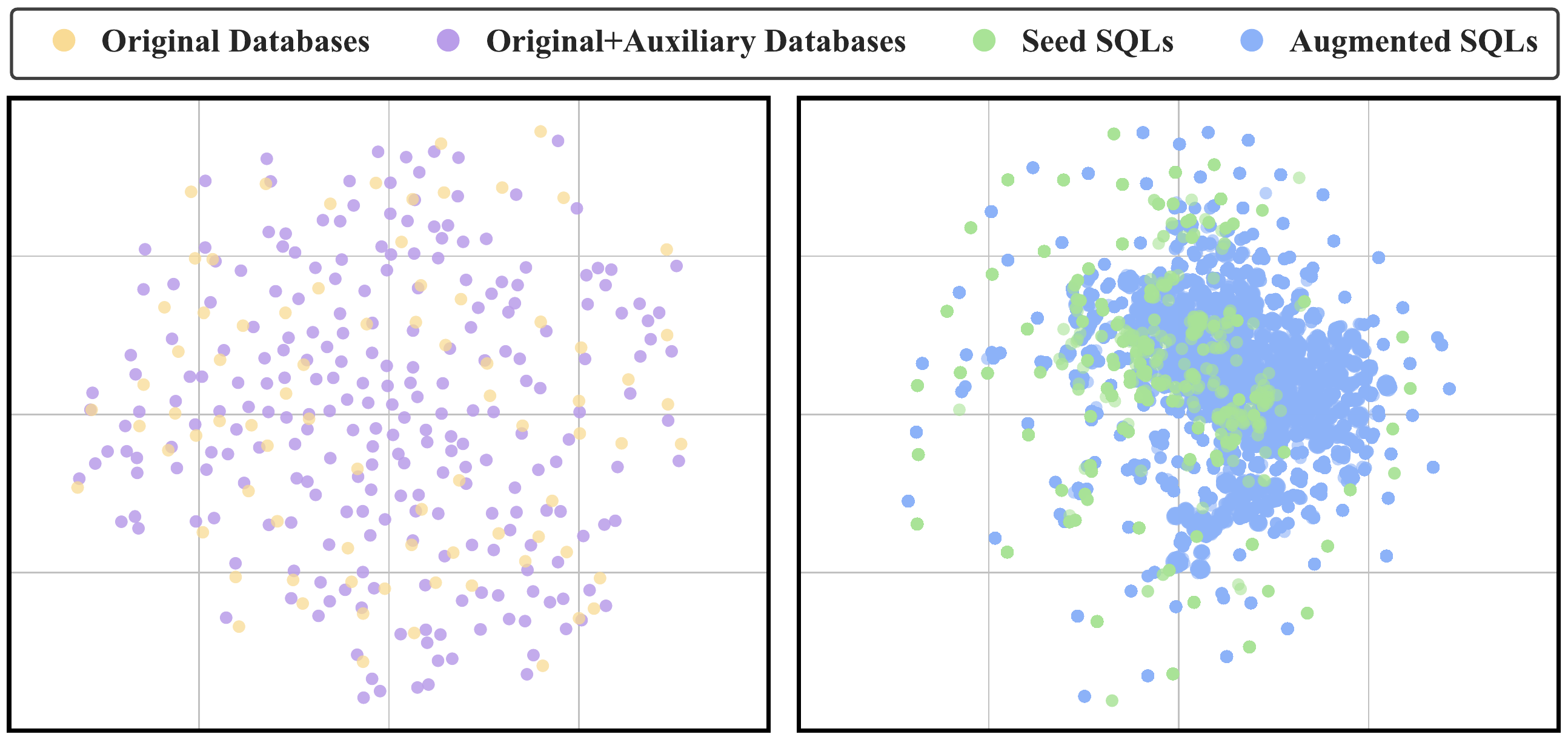}
\caption{Visualization of changes in distribution-level (left) and query-level (right) diversity induced by data augmentation.}
\label{fig:sql_comparision}
\end{figure}

\subsubsection{\textbf{Augmented Data Overview}}
\label{sec:augmented_data_overview}
Leveraging the proposed Text-to-SQL data augmentation pipeline, we generate an augmented dataset from the original SQL-query pairs \( s_i^{\text{ori}} \). The pipeline systematically synthesizes diverse and valid Text-to-SQL examples, which can serve as a valuable supplement to existing Text-to-SQL datasets. The final dataset comprises \( N \) entries of the following form:
\begin{align*}
    \left\{ s_i^{\text{aug}},\, q_i,\, S,\, p_i,\, \text{cot}_i\right\}_{i=1}^{N}
\end{align*}

 Using \textsc{Text2SQL-Flow}, we construct a dataset named \textsc{SQLFlow}. The raw seed data is sourced from widely used benchmarks. The final dataset comprises a total of 75,386 entries: 30,502 derived from the Spider-train dataset~\cite{yu2018spider}, 33,163 from the BIRD-train dataset~\cite{li2024can}, and 11,721 from the EHRSQL-train dataset~\cite{lee2022ehrsql}. In this paper, we refer to these subsets as SQLFlow-Spider, SQLFlow-BIRD, and SQLFlow-EHRSQL, respectively. \revision{It should be noted that the augmented data do not overlap with the original training datasets.
In addition, an \texttt{n-gram}-based filtering strategy is applied to ensure that \textsc{SQLFlow} has no overlap with the test data, thereby preventing any potential data leakage.}
Overall, our approach augments existing public datasets and is scalable, efficient, and capable of generating high-quality synthetic data.
 As shown in Table~\ref{tab:overall_statistics}, our pipeline not only mitigates the sparsity and structural gaps inherent in the original dataset but also substantially extends its complexity frontier—surpassing the original in both scale and quality. \revision{The broader coverage and increased diversity enabled by data augmentation are reflected in Figure~\ref{fig:sql_comparision}.
Using a 2D t-SNE projection, the figure illustrates both database-level and feature-level distributions.
The left panel indicates that auxiliary databases expand the coverage beyond the original dataset, while the right panel reveals a more diverse spread in feature representations derived from the statistics in Table~\ref{tab:overall_statistics}.}

\subsection{Database Interaction Implementation}
\label{sec:database_manager}
\revision{Multiple stages of the \textsc{Text2SQL-Flow} data augmentation pipeline require frequent interactions with the underlying database, including executing SQL queries for validity checking and accessing schema information for query generation. To support multiple database systems and improve implementation robustness, we implement a database interaction layer that abstracts database-specific driver details and provides a unified interface to upper-layer components. This interface supports batched SQL execution and schema access through standardized APIs, enabling seamless integration of different relational database systems (e.g., SQLite, MySQL) without modifying the core pipeline logic. For efficiency, schema information is cached after the first access to avoid redundant metadata queries during large-scale SQL generation.} 


\subsection{Synthetic Data for Open-Source Models}
\label{open_source_model}


The pairs \( \langle p_i, cot_i \rangle \) generated above serve as high-quality training data $\mathcal{D}$. 
During SFT, the model is trained on the full output sequence using standard causal language modeling:
\begin{equation}
\mathcal{L} = -\mathbb{E}_{(x, y) \sim \mathcal{D}} \left[ \sum_{t=1}^{|y|} \log P_{\theta}(y_t \mid x, y_{<t}) \right]
\end{equation}
where the input $x$ corresponds to the prompt $p_i$, and the output $y$ represents the complete CoT sequence $cot_i$. 
The model parameters are denoted by $\theta$. 
By optimizing this loss over the entire output sequence, the model is encouraged to learn interpretable reasoning pathways and improve its performance on the Text-to-SQL task.

\subsection{Synthetic Data for Prompting-based Methods}
\label{close_source_model}

For closed-source LLMs, whose parameters are not accessible, prompt-based methods have become a crucial technique for improving performance. A typical prompt provided to an LLM generally includes multiple components: an instruction, database schema information, and the NL question posed by the user. To enhance the model's generalization ability across different questions, few-shot examples are often retrieved and incorporated into the prompt context in practice. These examples are typically selected due to their similarity to the target question, serving as guidance for the model in generating the corresponding SQL query. Each few-shot example consists of an NL question paired with its corresponding SQL query. Ideally, such examples should effectively reveal the structural patterns and generation logic of the target SQL, thereby improving the model's reasoning performance.

Two problems need to be considered during few-shot example retrieval:  
\textit{(1) What information is available during retrieval?} During retrieval, only the user’s NL question is available, and we cannot know the ground-truth SQL query in advance. Moreover, any intermediate representation derived from this NL question may introduce information loss, thereby degrading retrieval quality.  
\textit{(2) Which kind of examples do we need?} The primary value of few-shot examples lies in providing structural guidance for SQL generation. Therefore, the SQL queries of the retrieved examples should exhibit high similarity to the target query, but not necessarily the higher the similarity, the better. Notably, different components of the NL question and SQL query contribute unequally to the similarity. Specifically, the ``skeleton'' of a question reflects its basic querying pattern, while the skeleton of an SQL query manifests as the logical composition of core clauses such as ``SELECT–FROM–WHERE–JOIN''. These skeletal elements encode the query’s logical intent and functional structure, serving as critical signals for guiding model generation. In contrast, fine-grained details, such as specific table names, column names, string literals, or numeric constants, are highly dependent on the particular database schema and application context. If directly used in similarity computation, they often introduce noise and impair retrieval effectiveness. Thus, an ideal retrieval mechanism should identify examples whose corresponding SQL queries are structurally aligned with the target query, based solely on the NL question.

\label{knowledge_base}
However, existing methods for retrieving few-shot examples fail to provide effective guidance because similarity has not been sufficiently exploited. In this case, we propose a structure-aware few-shot retrieval method for Text-to-SQL. The core idea is to map both NL questions and SQL queries into a unified, detail-abstracted semantic space, where retrieval is performed based on structural similarity. Specifically, inspired by the masking strategy in DAIL--SQL~\cite{gao2023text}, we apply standardized masking to schema-dependent elements (e.g., table/column names), string literals, and numeric constants in both questions and SQL queries, denoted as $\text{Mask}(q)$ and $\text{Mask}(\text{s})$, respectively.
Building upon this, we develop a structure-aware embedding model to compute semantic similarity between masked NL questions and masked SQL queries.
\revision{Given a target question $q_t$, we retrieve the top-$k$ examples from a retrieval pool $B = \{(q_i, s_i)\}_{i=1}^N$, which we refer to as a knowledge base.
$B$ consists of NL--SQL pairs from the training dataset, with no overlap with the test dataset.
$B$ is stored as an unordered collection of independent data pairs, rather than a structured knowledge graph.}

We fine-tune our embedding model using SFT and employ a contrastive learning framework for training. The training data consist of numerous $(q_i, s_i)$ pairs. For each sample, the NL question $q_i$ serves as the query, its corresponding SQL $s_i$ as the positive example, and $k$ randomly sampled SQL queries from the knowledge base (excluding $s_i$) as negative examples. The model is trained by optimizing the InfoNCE loss:
\begin{equation}
\label{eq:embedding_loss}
\mathcal{L} = -\log \frac{\exp(\text{sim}(q, s^+) / \tau)}{\exp(\text{sim}(q, s^+) / \tau) + \sum_{j} \exp(\text{sim}(q, s_j^-) / \tau)}
\end{equation}
where $\mathbf{e}_q^i = \text{Embed}(\text{Mask}(q_i))$ and $\mathbf{e}_s^i = \text{Embed}(\text{Mask}(s_i))$ denote the embedding vectors of the masked question and its corresponding SQL, respectively; $\text{sim}(\cdot, \cdot)$ is the cosine similarity function; and $\tau$ is a temperature parameter.

In our implementation, we adopt the Qwen3-Embedding architecture~\cite{zhang2025qwen3}, employing a causal-attention-based LLM as the embedding encoder. Both the knowledge base and the training data for the embedding model are derived from the \textsc{SQLFlow} dataset we constructed. The input format concatenates the task instruction with the masked question as follows:  
$\texttt{\{Instruction\} \{Mask}(q_i)\texttt{\}}$
followed by an [EOS] token. The final sentence embedding is obtained from the hidden state of the [EOS] token in the last layer of the encoder.
Our method retrieves few-shot examples that are highly aligned with the target query in SQL structure. This enhances the few-shot reasoning capability of closed-source LLMs on the Text-to-SQL task.

\section{Experiments}\label{sec:exp}
\subsection{Experimental Setup}

\subsubsection{Benchmarks}
Multiple benchmarks are used to evaluate the effectiveness of our method. 

\begin{itemize}
    \item \textbf{Spider}~\cite{yu2018spider} contains 10,181 NL questions and 5,693 unique SQL queries, involving 200 databases across multiple tables covering 138 different domains. We utilize both its development set, containing 1,034 samples, and its test set with 2,147 samples for evaluation. 
    \item \textbf{BIRD}~\cite{li2024can} includes 12,751 NL question-SQL pairs and 95 databases across 37 domains, focusing on large-scale noisy data and external knowledge reasoning. Our experiments are performed on its development set, which consists of 1,534 data samples.
    \item \textbf{EHRSQL}~\cite{lee2022ehrsql} includes approximately 24,000 NL question-SQL pairs linked to 2 open-source electronic health record databases. The dataset introduces challenges such as complex, time-sensitive questions and the inclusion of unanswerable questions.
    \item \textbf{Spider-DK, Spider-Syn, and Spider-Realistic}~\cite{gan2021exploring, gan2021towards, deng2020structure} are three widely-adopted robustness benchmarks designed to evaluate model performance in practical Text-to-SQL applications. Spider-DK tests the model's ability to understand implicit domain knowledge in NL questions, offering 535 samples for evaluation. Spider-Syn replaces schema-related words with manually selected synonyms and provides 1,034 evaluation samples. Spider-Realistic removes or paraphrases explicit mentions of column names, offering 508 samples for evaluation.
\end{itemize}

\begin{table*}[t]
\centering
\caption{Performance of LLMs on mainstream benchmarks. 
\revision{The first two blocks list closed-source and open-source base models, respectively, which are included as reference baselines.
The last two blocks present fine-tuned models, where the first column indicates the training data setting.}``Seed Data'' denotes the union of Spider-train, BIRD-train, and EHRSQL-train.}

\label{tab:llm_results}
\resizebox{\textwidth}{!}{
\begin{tabular}{l|cccccccccccccc|cc}
\toprule
\multirow{3}{*}{\textbf{LLM / Training Data}} & \multicolumn{2}{c}{\textbf{Spider}} & \multicolumn{2}{c}{\textbf{Spider}} & \multicolumn{2}{c}{\textbf{BIRD}} & \multicolumn{2}{c}{\multirow{2}{*}{\textbf{EHRSQL}}} & \multicolumn{2}{c}{\textbf{Spider-}} & \multicolumn{2}{c}{\textbf{Spider-}} & \multicolumn{2}{c|}{\textbf{Spider-}} & \multicolumn{2}{c}{\multirow{2}{*}{\textbf{Average}}}\\

& \multicolumn{2}{c}{\textbf{dev}} & \multicolumn{2}{c}{\textbf{test}} & \multicolumn{2}{c}{\textbf{dev}} & & & \multicolumn{2}{c}{\textbf{DK}}& \multicolumn{2}{c}{\textbf{Syn}} & \multicolumn{2}{c|}{\textbf{Realistic}} & &\\

\cmidrule(lr){2-3}\cmidrule(lr){4-5}\cmidrule(lr){6-7}
\cmidrule(lr){8-9}\cmidrule(lr){10-11}\cmidrule(lr){12-13}\cmidrule(lr){14-15}\cmidrule(lr){16-17}
& Gre & Maj & Gre & Maj & Gre & Maj & Gre & Maj & Gre & Maj & Gre & Maj & Gre & Maj &Gre & Maj\\

\midrule
\rowcolor[rgb]{.867, .922, .969}
\multicolumn{17}{c}{\textit{\textbf{Closed-source LLMs}}} \\
\midrule
GPT-4-Turbo & 72.4 & 72.2 & 83.4 & 84.2 & 62.0 & 63.6 & 43.1 & 44.8 & 72.3 & 72.1 & 62.9 & 63.5 & 67.5 & 68.3 & 66.2 & 67.0\\
GPT-4o & 70.9 & 70.7 & 83.2 & 84.9 & 61.9 & 64.0 & 44.9 & 45.5 & 72.9 & 73.5 & 59.6 & 62.3 & 66.5 & 66.7 & 65.7 & 66.8 \\
\midrule
\rowcolor[rgb]{.867, .922, .969}
\multicolumn{17}{c}{\textit{\textbf{Open-source LLMs}}} \\
\midrule
DeepSeek-Coder-7B-Instruct & 63.2 & 63.2 & 70.5 & 73.2 & 43.1 & 48.0 & 28.6 & 33.9 & 60.9 & 64.1 & 49.9 & 51.7 & 58.7 & 58.9 & 53.6 & 56.1 \\
Qwen2.5-Coder-7B-Instruct & 73.4 & 77.1 & 82.2 & 85.6 & 50.9 & 61.3 & 24.3 & 36.9 & 67.5 & 73.6 & 63.1 & 66.9 & 66.7 & 70.5 & 61.2 & 67.4\\
OpenCoder-8B-Instruct & 59.5 & 59.5 & 68.3 & 70.1 & 37.5 & 45.3 & 21.9 & 29.9 & 62.6 & 64.7 & 46.0 & 46.1 & 49.0 & 49.4 & 49.3 & 52.1 \\
Meta-Llama-3.1-8B-Instruct & 61.8 & 67.7 & 72.2 & 78.5 & 42.0 & 53.1 & 24.6 & 33.7 & 62.6 & 69.9 & 53.1 & 59.3 & 57.5 & 61.0 & 53.4 & 60.5 \\
Granite-8B-Code-Instruct & 58.5 & 59.2 & 64.9 & 68.6 & 27.6 & 32.5 & 16.0 & 22.6 & 50.7 & 54.4 & 45.0 & 46.8 & 48.8 & 49.4 & 44.5 & 47.6 \\
\midrule
\rowcolor[rgb]{.867, .922, .969}
\multicolumn{17}{c}{\textit{\textbf{Trained on Meta-Llama-3.1-8B-Instruct}}} \\
\midrule
SynSQL(75k) & 78.1 & 80.9 & 78.1 & 81.8 & 47.7 & 54.9 & 28.9 & 35.9 & 66.4 & 67.5 & 67.5 & 72.1 & 74.0 & 77.0 & 63.0 & 67.2 \\
SynSQL(2.5M) & 81.2 & 81.6 & 87.9 & 88.3 & 63.9 & 66.1 & 34.9 & 40.0 & 76.1 & 77.8 & 69.7 & 69.6 & 76.2 & 78.0 & 70.0 & 71.6 \\
\midrule
Seed Data + \textsc{SQLFlow} & 81.4 & 85.4 & 81.9 & 85.2 & 54.5 & 59.5 & 61.2 & 61.5 & 67.5 & 72.7 & 71.2 & 76.5 & 78.0 & 80.1 & 70.8 & 74.4\\
\textbf{\textsc{SQLFlow}} & 80.4 & 83.3 & 81.3 & 83.9 & 51.0 & 57.1 & 51.2 & 56.0 & 66.0 & 72.0 & 69.3 & 74.0 & 75.4 & 79.7 & 67.8 & 72.3 \\

\midrule
\rowcolor[rgb]{.867, .922, .969}
\multicolumn{17}{c}{\textit{\textbf{Trained on Qwen2.5-Coder-7B-Instruct}}} \\
\midrule
SynSQL(75k) & 80.7 & 83.6 & 81.6 & 83.7 & 53.2 & 59.8 & 34.6 & 43.8 & 68.4 & 67.9 & 68.8 & 71.9 & 75.8 & 79.5 & 66.2 & 70.0 \\
SynSQL(2.5M) & 81.2 & 81.6 & 87.9 & 88.3 & 63.9 & 66.1 & 34.9 & 40.0 & 76.1 & 77.8 & 69.7 & 69.6 & 76.2 & 78.0 & 70.0 & 71.6 \\
\midrule
Seed Data + \textsc{SQLFlow} & 84.9 & 86.2 & 85.3 & 87.3 & 58.0 & 62.1 & 61.2 & 62.0 & 74.4 & 75.9 & 74.8 & 78.0 & 82.1 & 83.5 & 74.4 & 76.4\\
\textbf{\textsc{SQLFlow}} & 83.4 & 85.0 & 85.1 & 86.6 & 56.8 & 61.0 & 53.7 & 57.5 & 73.3 & 74.8 & 73.4 & 76.4 & 81.9 & 81.7 & 72.5 & 74.7 \\
\bottomrule
\end{tabular}%
}
\end{table*}

\subsubsection{Evaluation Metric}
Execution accuracy (EX) is adopted in our experiments. This metric compares the execution result of the predicted SQL query against that of the ground-truth SQL query on a live database instance.

\label{baseline_of_LLM_training}
\subsubsection{Baselines of LLM Training}
\revision{Multiple open-source and closed-source mainstream LLMs are evaluated as reference baselines.
The closed-source models include GPT-4-Turbo~\cite{achiam2023gpt} and GPT-4o~\cite{hurst2024gpt}.
The open-source models encompass DeepSeek-Coder-7B-Instruct~\cite{guo2024deepseek}, Qwen2.5-Coder-7B-Instruct~\cite{hui2024qwen2}, OpenCoder-8B-Instruct~\cite{huang2024opencoder}, Meta-Llama-3.1-8B-Instruct~\cite{grattafiori2024llama}, and Granite-8B-Code-Instruct~\cite{mishra2024granite}.}
Regarding the training data for the models, we utilize not only the public datasets Spider-train~\cite{yu2018spider} and BIRD-train~\cite{li2024can}, but also the synthetic dataset SynSQL proposed by OmniSQL~\cite{li2025omnisql}.
We randomly sample a subset of SynSQL containing 75K examples (75,386 instances), matching the exact size of the \textsc{SQLFlow} dataset for ease of comparison.
In addition, we use the full SynSQL-2.5M dataset, comprising 2.5 million examples, as a reference.
\revision{We use GPT-4o as the backend model to construct \textsc{SQLFlow} and set the temperature to zero.}

\subsubsection{Baselines of Few-shot Example Retrieval}\label{sec:baselines}
\revision{
We compare our retrieval method with multiple example retrieval strategies under both masked and unmasked settings.
The masking strategy in DAIL-SQL~\cite{gao2023text} is applied here.
}
All similarity calculations use cosine similarity. Strategies are as follows:
\begin{itemize}
\item \textbf{Zero-shot}~\cite{kojima2022large} refers to a setting in which no few-shot examples are used during inference.
\item \textbf{Random Selection} means selecting the examples from the knowledge base randomly without any criteria.
\item \textbf{Question Similarity Selection}~\cite{liu2023comprehensive} retrieves examples by selecting candidate questions with the highest similarity to the target question.
\item \textbf{DAIL Selection}~\cite{gao2023text} retrieves examples by jointly considering question- and query-level similarity. Domain-specific tokens in questions are first masked, after which candidates are ranked by question similarity and filtered based on whether their query Jaccard similarity to a pre-generated reference SQL exceeds a predefined threshold.
\item \revision{\textbf{Question-SQL Similarity Selection} encodes the target question and each candidate SQL into a shared embedding space with the untrained model, and selects the examples with the highest embedding similarity.}
\item \revision{\textbf{PreSQL-SQL Similarity Selection} ranks examples by the embedding similarity between each candidate SQL and a pre-generated SQL for the target question, and selects the most similar ones.}
\item \revision{\textbf{Question-AST Similarity Selection} parses each candidate SQL into an abstract syntax tree (AST) with \texttt{SQLGlot} and converts it into a compact, structure-preserving representation, which keeps all child expressions to retain query structure, while pruning semantically redundant fields (e.g., None arguments and default flags). Candidates are then ranked by the embedding similarity between the target question and the AST representation.}
\end{itemize}

\newcolumntype{Y}{>{\centering\arraybackslash}X}
\begin{table*}[t]
  \centering
  \caption{Performance comparison of retrieval strategies. ``Upper limit'': DAIL selection retrieval using ground-truth SQL. ``API'' indicates whether LLM API calls are required during retrieval, with an average token cost of 517 tokens per query.}
  \label{tab:retrieval_performance}
  \begin{tabularx}{\textwidth}{l|Y|YYYYYY|Y}
    \toprule
    \multirow{2}{*}{\textbf{Retrieval Strategy}} & \multirow{2}{*}{\textbf{API}} & {\textbf{BIRD}} & {\textbf{Spider}} & {\textbf{Spider}} & {\textbf{Spider-}} & {\textbf{Spider-}} & {\textbf{Spider-}} & \multirow{2}{*}{\textbf{Average}}\\
     &  & {\textbf{dev}} & {\textbf{dev}} & {\textbf{test}} & {\textbf{DK}} & {\textbf{Realistic}} & {\textbf{Syn}} & \\

    \midrule
    \rowcolor[rgb]{.867, .922, .969}
    \multicolumn{9}{c}{\textit{\textbf{Reference strategies without mask}}} \\
    \midrule
    Zero-shot & - & 52.5 & 74.1 & 75.5 & 60.2 & 71.1 & 62.2 & 65.9 \\
    Random selection & \ding{56} & 56.6 & 78.3 & 79.8 & 64.5 & 72.2 & 66.6 & 69.7\\
    \midrule
    Question similarity & \ding{56} & 58.3 & 79.6 & 81.9 & 66.2 & 74.4 & 67.7 & 71.4\\
    \midrule
    DAIL selection & \ding{52} & 58.8 & 80.0 & \underline{82.2} & 66.4 & 75.2 & 68.9 & 71.9\\
    \midrule
    PreSQL-SQL similarity & \ding{52} & 59.1 & 78.9 & 81.0 & 63.7 & 74.2 & 69.6 & 71.1\\
    \midrule
    Question-AST similarity & \ding{56} & 58.2 & 79.6 & 80.7 & 64.5 & 73.6 & 65.7 & 70.4\\
    \midrule
    Question-SQL similarity & \ding{56} & 58.7 & 80.9 & 81.7 & 65.6 & 74.2 & 66.4 & 71.3\\
    \midrule
    Upper limit & \ding{52} & \textbf{61.8} & \textbf{83.9} & \textbf{85.7} & \textbf{70.8} & \textbf{77.4} & \textbf{72.7} & \textbf{75.4}\\
    \midrule
    \textbf{SQLFlow-part+Spider+BIRD AST} & \ding{56} & 59.1 & 80.7 & 81.9 & 66.4 & 75.4 & 67.7 & 71.9\\
    \midrule
    \textbf{SQLFlow-part+Spider+BIRD SQL} & \ding{56} & \underline{59.3} & \underline{81.0} & 82.1 & \underline{67.3} & \underline{75.6} & \underline{70.1} & \underline{72.6}\\
    \midrule
    \rowcolor[rgb]{.867, .922, .969}
    \multicolumn{9}{c}{\textit{\textbf{Reference strategies with mask}}} \\
    \midrule
    Question similarity & \ding{56} & 59.0 & 81.0 & 83.1 & 67.5 & 75.6 & 69.9 & 72.7\\
    \midrule
    DAIL selection & \ding{52} & 59.4 & 80.8 & 82.9 & 67.1 & 75.2 & 69.8 & 72.5\\
    \midrule
    PreSQL-SQL similarity & \ding{52} & 59.3 & 77.5 & 78.5 & 63.6 & 71.9 & 67.5 & 69.7\\
    \midrule
    Question-AST similarity & \ding{56} & 56.8 & 78.2 & 79.6 & 65.4 & 73.0 & 67.2 & 70.0\\
    \midrule
    Question-SQL similarity & \ding{56} & 57.8 & 80.1 & 80.3 & 65.6 & 73.4 & 69.0 & 71.0\\
    \midrule
    Upper limit & \ding{52} & \textbf{61.6} & \textbf{83.8} & \textbf{85.1} & \textbf{69.2} & \textbf{78.7} & \textbf{73.6} & \textbf{75.3}\\
    \midrule
    \textbf{SQLFlow-part+Spider+BIRD AST} & \ding{56} & 59.5 & \underline{81.3} & 83.0 & \underline{68.8} & 76.4 & 70.9 & 73.3\\
    \midrule
    \textbf{SQLFlow-part+Spider+BIRD SQL} & \ding{56} & \underline{61.0} & 81.2 & \underline{83.5} & 67.7 & \underline{77.2} & \underline{71.1} & \underline{73.6}\\
    \bottomrule
  \end{tabularx}
\end{table*}

\subsubsection{Implementation Details}
To validate the effectiveness of our framework, we conduct full-parameter SFT on two widely adopted open-source LLMs: Meta-Llama-3.1-8B-Instruct~\cite{grattafiori2024llama} and Qwen2.5-Coder-7B-Instruct~\cite{hui2024qwen2}. We use the entire \textsc{SQLFlow} dataset as training data. The training configuration includes a learning rate of \(2 \times 10^{-5}\), 2 training epochs, and a warmup ratio of 15\%. 
During LLM inference, we explore two decoding strategies: greedy decoding and majority voting. 
Greedy decoding (denoted as \texttt{Gre} in tables) uses a temperature of 0 to produce deterministic outputs. 
Majority voting (denoted as \texttt{Maj} in tables) samples 8 candidate responses per input at a temperature of 0.8. 
From these candidates, valid SQL queries are extracted and executed; the final prediction is selected as the query whose execution result receives the most votes.

As for the retrieval module, we fine-tune a Qwen3-Embedding-0.6B model~\cite{zhang2025qwen3} via SFT. 
To construct the training data, we use a filtered subset of the SQLFlow-Spider and SQLFlow-BIRD datasets (described in Section~\ref{sec:augmented_data_overview}), retaining only SQL queries that begin with a \texttt{SELECT} clause. 
This process yields 67{,}570 training instances, which we refer to as \textsc{SQLFlow-part}.
The embedding model is trained for three epochs with a learning rate of \(6 \times 10^{-6}\) and a per-device batch size of 2. 
During training, the NL question is treated as the query, and its corresponding SQL query is used as the positive example. 
For each query, 30 negative SQL examples are randomly sampled from the remaining dataset.
At inference time, we compute question--SQL similarity using cosine similarity between embeddings generated by both the original and the fine-tuned embedding models. 
We apply the masking strategy proposed in DAIL--SQL~\cite{gao2023text} during both training and inference. 
Retrieved examples are incorporated into the prompt using a 5-shot setting for LLM inference.
For experiments involving closed-source LLMs, we use GPT-4o with greedy decoding and set the temperature to zero. 
Prompt construction follows the organization strategy introduced in DAIL.
In addition to \textsc{SQLFlow-part}, we construct two baseline datasets for comparison. 
The first baseline combines the standard Spider-train and BIRD-train splits. 
The second baseline, denoted as \textsc{SynSQL-part}, consists of 67{,}570 SQL queries randomly sampled from SynSQL-2.5M~\cite{li2025omnisql}, matched in size with \textsc{SQLFlow-part} to ensure a fair comparison.

All experiments are conducted on an Ubuntu 22.04.4 LTS server equipped with eight NVIDIA H200 GPUs. LLM SFT is implemented using Llama-Factory v0.9.3, while retrieval model SFT is carried out with SWIFT v3.6.3. Inference for both models is handled by vLLM v0.10.0.

\subsection{Performance of LLM Fine-tuning}

As shown in Table~\ref{tab:llm_results}, the data generated by our framework leads to consistent performance improvements across multiple mainstream benchmarks, demonstrating the effectiveness of \textsc{Text2SQL-Flow}. We fine-tune two widely used open-source models, Meta-Llama-3.1-8B-Instruct and Qwen2.5-Coder-7B-Instruct, to evaluate the general applicability of our data. In both cases, models trained on our generated data significantly outperformed their respective baselines as well as other models. 
When fine-tuned with \textsc{SQLFlow} data, Qwen2.5-Coder-7B-Instruct achieved notable gains: execution accuracy (``Gre'') on Spider-dev increased from 80.7\% to 83.4\% (+2.7\%), on BIRD-dev from 53.2\% to 56.8\% (+3.6\%), and on the challenging EHRSQL benchmark from 34.6\% to 53.7\% (+19.1\%). These results confirm two key points: (1) SFT substantially enhances model performance on Text-to-SQL tasks, and (2) the data generated via our \textsc{Text2SQL-Flow} framework exhibits high quality and strong training utility. The consistent improvements across two distinct model architectures demonstrate the broad effectiveness of our dataset in boosting model performance.

In comparison with other training datasets, our data demonstrates clear advantages. 
Specifically, models trained with \textsc{SQLFlow} (75K) not only outperform the baseline models but also consistently surpass SynSQL (75K), highlighting the robustness of our data generation approach.
Remarkably, despite being trained on a substantially smaller dataset, the model fine-tuned with \textsc{SQLFlow} (75K) achieves performance comparable to SynSQL-2.5M on several challenging benchmarks.
Furthermore, when real-world seed data, including Spider-train, BIRD-train, and EHRSQL-train, are incorporated into the training set, the resulting model attains state-of-the-art (SOTA) performance across multiple benchmarks. These results indicate that expanding seed data with \textsc{SQLFlow} effectively enhances model performance.


\begin{table}[ht]
\centering
\caption{Comparison of training data constructed under different ablation settings.
\# Tokens per SQL indicates the average tokens consumed to generate each augmented SQL.}
\resizebox{\columnwidth}{!}{
\begin{tabular}{l|ccc|c}
\toprule
\multirow{2}{*}{\textbf{Training Setting}} & {\textbf{Spider}} & {\textbf{Spider}} & {\textbf{BIRD}} & {\textbf{\# Token}}\\
& {\textbf{dev}} & {\textbf{test}} & {\textbf{dev}} & {\textbf{per SQL}}\\
\midrule
Baseline & 78.3 & 79.8 & 52.5 & - \\
\midrule
Full Pipeline & 81.7 & 83.2 & 55.5 & 4909\\
\midrule
\rowcolor[rgb]{.867, .922, .969}
\multicolumn{5}{c}{\textit{\textbf{Auxiliary Database Selection}}} \\
\midrule
w/o Auxiliary Database & 81.1 & 82.3 & 54.8 & \multirow{2}{*}{0}\\
Random Selection & 80.9 & 81.7 & 54.0 & \\
\midrule
\rowcolor[rgb]{.867, .922, .969}
\multicolumn{5}{c}{\textit{\textbf{SQL Augmentation Strategy}}} \\
\midrule
w/o DVT & 80.9 & 82.5 & 54.0 & \multirow{6}{*}{857}\\
w/o QSM & 80.1 & 81.8 & 53.4 & \\
w/o BLC & 79.7 & 82.3 & 54.0 & \\
w/o CE & 79.7 & 81.9 & 53.8 & \\
w/o ASF & 81.1 & 81.5 & 55.5 & \\
w/o PO & 79.9 & 81.7 & 54.2 & \\
\midrule
w/o Executability Filter & 80.1 & 82.3 & 54.1 & 0\\
\midrule
\rowcolor[rgb]{.867, .922, .969}
\multicolumn{5}{c}{\textit{\textbf{Question Style}}} \\
\midrule
w/o TF & 80.4 & 83.1 & 53.5 & \multirow{2}{*}{1742}\\
w/o SSI & 80.1 & 82.0 & 55.4 & \\
w/o IDC & 80.9 & 82.3 & 54.0 & \\
w/o IP & 81.5 & 82.8 & 54.0 & \\
\midrule
w/o Correspondence Filter & 80.5 & 80.0 & 53.4 & 709\\
\midrule
w/o Prompt Generation & - & - & - & 0\\
\midrule
w/o Chain-of-Thought & 75.5 & 78.1 & 53.3 & 1601\\
\bottomrule
\end{tabular}
}
\label{tab:pipeline_ablation}
\end{table}

\subsection{Ablation Study of the Text2SQL-Flow Pipeline}
\label{sec:ablation}

\revision{
We conduct an ablation study to analyze the contribution of each component in \textsc{Text2SQL-Flow}. Results are reported in Table~\ref{tab:pipeline_ablation}.
We select 20 databases from Spider as original databases and 20 databases from BIRD as auxiliary databases, from which 200 representative SQLs are sampled per source using SQL-structure-based bucketing, resulting in 400 seed SQLs. Each seed SQL is augmented using all six SQL augmentation strategies, each instantiated twice, yielding $4,800$ augmented SQLs. Each augmented SQL is paired with four question styles, producing $19{,}200$ SQL--question pairs. All components of our pipeline are enabled unless explicitly ablated.
The Baseline (line~1) is trained solely on the seed SQLs, whereas all other settings are trained on augmented data with a fixed size of 2{,}400 instances.
The Full Pipeline (line~2) samples training data evenly from Spider and BIRD.
w/o Auxiliary Database (line~3) draws samples exclusively from the Spider source while keeping the total number of sampled instances fixed at 2{,}400, whereas Random Database Selection (line~4) replaces the structured auxiliary database selection strategy with random sampling.
For SQL augmentation and question-style ablations, one strategy or style is removed at a time, with uniform sampling over the remaining ones; the corresponding category names follow the abbreviations defined in Section~\ref{sql_augmentation} and Section~\ref{question_generation}.
In w/o Executability Filter (line~11) and w/o Correspondence Filter (line~16), invalid or misaligned samples are retained during training. w/o Chain-of-Thought (line~18) removes the CoT rationales while keeping the same training samples.
}

\revision{Full Pipeline consistently outperforms the baseline, demonstrating the effectiveness of \textsc{Text2SQL-Flow}. Removing auxiliary databases degrades performance, while random database selection further harms results, validating the necessity of auxiliary databases and our structured coverage strategy. Ablating any SQL augmentation strategy or question style leads to performance drops, indicating that both SQL-level diversity and linguistic variation are important. Disabling executability or alignment filters consistently hurts performance, highlighting the importance of data quality control. Removing CoT supervision causes the most severe degradation, underscoring the critical role of explicit reasoning signals.}

\begin{table}[h]
\centering
\caption{
Performance of models trained on the same amount of synthetic data generated by different LLMs. ``API cost'' is reported in USD per million tokens and is set to zero for open-source models, excluding local compute overhead.}
\label{tab:llm_robustness}
\resizebox{\columnwidth}{!}{
\begin{tabular}{l|c|ccc}
\toprule
\textbf{LLM Backbone} 
& \textbf{API Cost} 
& \textbf{Spider dev} 
& \textbf{Spider test} 
& \textbf{BIRD dev} \\
\midrule
Baseline & - & 78.7 & 78.0 & 50.3 \\
\midrule
\rowcolor[rgb]{.867, .922, .969}
\multicolumn{5}{c}{\textit{\textbf{Closed-source LLMs}}} \\
\midrule
Claude-3.7-Sonnet 
& 12 
& 80.8 
& 83.3 
& 50.9 \\
Gemini-2.5-Flash 
& 0.6 
& 82.8 
& 83.0 
& 52.3 \\
GPT-3.5-Turbo 
& 0.5 
& 80.9 
& 81.7 
& 51.6 \\
GPT-4o 
& 2.5 
& 81.2 
& 81.7 
& 54.1 \\
\midrule
\rowcolor[rgb]{.867, .922, .969}
\multicolumn{5}{c}{\textit{\textbf{Open-source LLMs}}} \\
\midrule
DeepSeek-V3 
& 0 
& 79.8 & 81.0 
& 50.7 \\
GLM-4.6 
& 0 
& 80.9 
& 81.0 
& 51.4\\
Qwen-3-235B 
& 0 
& 80.5 
& 80.6 
& 50.8 \\
\bottomrule
\end{tabular}
}
\end{table}

\subsection{Robustness to the Choice of Data Generation Models}
\label{sec:llm_robustness}

\revision{Many components of \textsc{Text2SQL-Flow} rely on LLMs, making the choice of the data generation backbone a potentially influential factor for the quality of the augmented data.
To examine this sensitivity, we instantiate the same SQL data augmentation pipeline with a diverse set of both closed-source and open-source LLMs. Throughout this study, we keep the overall pipeline design, augmentation procedures, and the scale of synthetic data fixed, and vary only the LLM used for data generation.
Specifically, each model is prompted to generate 1,000 synthetic training examples based on 500 samples randomly selected from the Spider training set. The resulting augmented data are then used to train the downstream Text-to-SQL model.
Baseline (line~1) refers to the model trained using data constructed solely from these 500 samples.}

\revision{As shown in Table~\ref{tab:llm_robustness}, the downstream performance remains competitive across a wide range of data generation models. While stronger LLMs tend to produce slightly better synthetic data, even relatively weaker or low-cost models, including GPT-3.5-Turbo and open-source LLMs, still yield solid performance on benchmarks. This suggests that \textsc{Text2SQL-Flow} does not critically depend on highly sophisticated LLM reasoning during data generation. Instead, its robustness primarily stems from the structured augmentation process, executability filtering, and semantic alignment, which together mitigate noise introduced by imperfect generation models. Across all LLM backbones, the performance variation remains within a moderate range, with all models outperforming the baseline, without altering the overall effectiveness of the pipeline.}

\subsection{Comparison of Retrieval Strategies}\label{sec:retrieval_analysis}

We compare our retrieval method with several baselines, where the embedding model is trained on SQLFlow-part, Spider-train, and BIRD-train, using Qwen3-Embedding-0.6B as the base model.
As shown in the lower part of Table~\ref{tab:retrieval_performance}, fine-tuning the retrieval model on high-quality synthetic data from SQLFlow-part using our retrieval strategy significantly boosts performance. Both the high-quality augmented data and the masked alignment retrieval method contribute to this improvement. Specifically, our Question-SQL alignment method without masking achieves 59.3\%, 81.0\%, and 82.1\% accuracy on BIRD-dev, Spider-dev, and Spider-test, respectively, outperforming the baseline that relies on Question-SQL similarity retrieval with the base (non-finetuned, no mask) model. This indicates that fine-tuning with augmented data enables the model to select more relevant examples, thereby enhancing the in-context learning capability of downstream closed-source models for accurate SQL generation.

The masking strategy is another key contributor to this improvement. Ablation studies confirm its effectiveness: for non-finetuned methods, masking consistently improves retrieval performance (e.g., DAIL selection accuracy on BIRD-dev increases from 58.8\% to 59.4\%). Similarly, when we fine-tune on SQLFlow-part and use masking during training, performance improves to 61.0\%, 81.2\%, and 83.5\% on the three benchmarks.
By masking literal values in the training data, the model learns to focus on the structural correspondence between questions and SQL queries while ignoring superficial lexical differences. This enables it to retrieve examples whose SQL patterns closely match the unseen ground-truth query based solely on the input question in a single step. \revision{We also evaluate ASTs as an alternative SQL representation for embedding-model training and example retrieval in our Question-AST similarity setting. However, converting SQL into ASTs can obscure semantic cues but may introduce additional noise, leading to fewer improvements on some benchmarks. Since SQL already encodes rich structural information, our Question-SQL alignment is designed to enable the retrieval model to learn question–query correspondences in a data-driven manner while preserving the full SQL signal.} In contrast, strategies such as DAIL Selection and PreSQL-SQL Selection require generating an intermediate SQL query for retrieval, which incurs additional API costs and risks propagating errors from intermediate generations.

\subsection{Comparison of Training Data for Retrieval Model}

\begin{figure}[t] 
\centering
\includegraphics[width=\columnwidth]{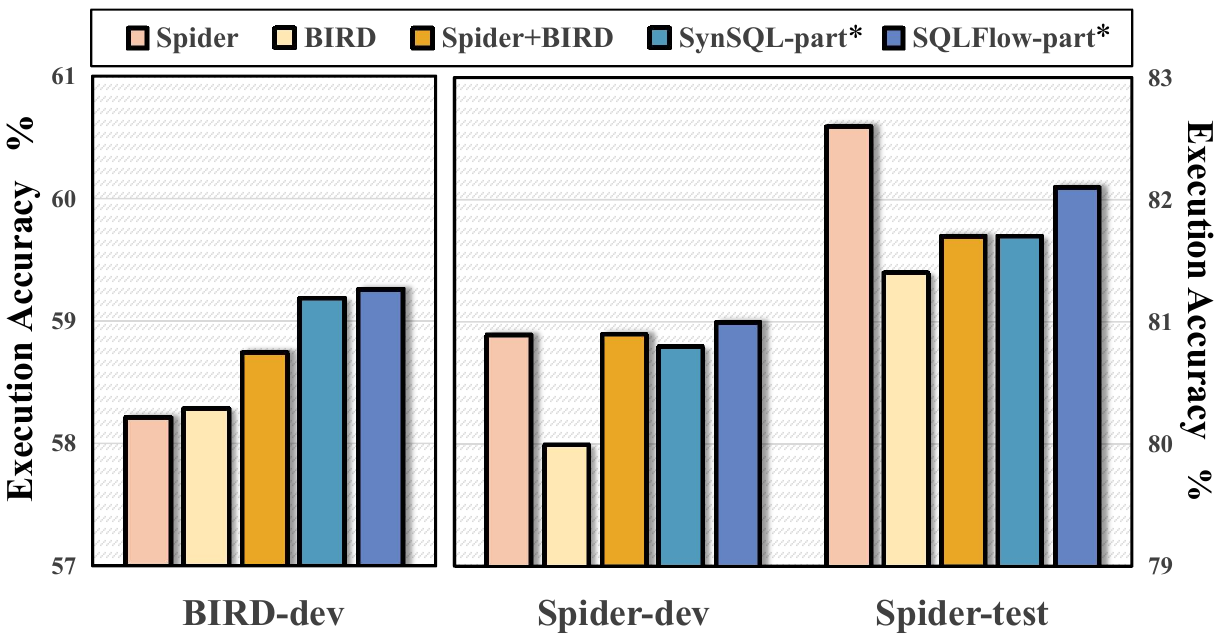}
\caption{Performance of question–SQL alignment strategy retrieval models trained on different datasets.
No masking operations are applied during the retrieval process. ``SynSQL-part*'' and ``SQLFlow-part*'' denote the corresponding datasets combined with Spider-train and BIRD-train.}
\label{fig:SFT_performance}
\end{figure}

As shown in Figure~\ref{fig:SFT_performance}, the model trained solely on the BIRD dataset performs well on BIRD-dev but suffers a sharp performance drop on Spider-dev and Spider-test, indicating limited cross-benchmark generalization when training on a single dataset.
However, simply mixing data from multiple sources does not necessarily enhance generalization. For example, training on both Spider and BIRD data for the Spider benchmark even leads to degraded performance. In contrast, the model trained on the SQLFlow-part combined with Spider and BIRD achieves better results, effectively mitigating overfitting and substantially improving cross-benchmark performance. This indicates that the benefit does not stem merely from larger data volume, but from the diversity and quality of the data generated by \textsc{Text2SQL-Flow}, which more effectively boosts model generalization.
A comparison with SynSQL-part further validates this conclusion. As a fully synthetic dataset, SynSQL exhibits a data distribution that diverges significantly from real-world applications. Although it aims for generalization, its effectiveness on specific tasks remains limited, demanding massive data and computation for only modest gains. For instance, SynSQL-part achieves only slight improvement on the Spider benchmark, barely exceeding the Spider+BIRD combination. In contrast, SQLFlow-part delivers markedly better results, highlighting its superior data quality, distribution alignment, and task relevance.

\section{Conclusion}

In this work, we introduced \textsc{Text2SQL-Flow}, a robust SQL-aware data augmentation framework that enables the scalable generation of high-quality, structurally diverse Text-to-SQL examples from limited seed data. 
Leveraging this framework, we constructed \textsc{SQLFlow}, a large-scale dataset comprising 75,386 annotated instances. Experimental results show that models fine-tuned on \textsc{SQLFlow} consistently outperform those trained on existing comparable datasets, highlighting the critical role of data quality and structural diversity in enhancing model generalization and compositional reasoning. 
Furthermore, our proposed masked alignment retrieval method improves the in-context learning performance of closed-source large language models by enabling structure-aware example selection. 
This work establishes a data-centric foundation for future Text-to-SQL research and demonstrates the importance of high-quality data in data-centric AI.


\newpage

\bibliographystyle{IEEEtran}
\bibliography{IEEEabrv,main_arxiv.bib}

%% file: contents/appendix.tex
\appendices

\section{Additional Pipeline Components}

In practical applications, to facilitate the organization and management of data generated by the pipeline and to support downstream usage, we additionally introduce two classifier steps to categorize the generated SQL queries. These operators are primarily user-oriented and are designed to improve system usability and engineering practicality, rather than to directly enhance the model performance emphasized in this paper. For this reason, their detailed descriptions are deferred to the Appendix.
Based on the resulting classifications, users can more conveniently and efficiently inspect, analyze, and utilize the generated data. We describe the two operators as follows.

\subsection{\textbf{SQL Component Difficulty Classifier}}
Classifying generated SQL queries facilitates a deeper analysis and understanding of their structural complexity. Following the evaluation criteria established in the Spider benchmark, we categorize SQL queries into four difficulty levels: \textit{easy}, \textit{medium}, \textit{hard}, and \textit{extra hard}. This classification is primarily based on the number and complexity of syntactic components present in the query, including column selections and the use of aggregate functions in the \texttt{SELECT} clause, as well as the presence of keywords such as \texttt{GROUP BY}, \texttt{ORDER BY}, \texttt{INTERSECT}, or advanced constructs like nested subqueries. Generally, queries incorporating a greater number of such components are considered structurally more complex and thus assigned a higher difficulty level. Under this principle, the augmented SQL query \( s_i^{\text{aug}} \) can be assigned a reasonable and quantifiable component-based difficulty. This process is formally expressed as:

\begin{equation}
    cd_i = \text{CC}(s_i^{\text{aug}})
\end{equation}

where \( \text{CC}(\cdot) \) denotes the mapping function of the classifier, and \( cd_i \) represents the component difficulty of the query.

\subsection{\textbf{SQL Execution Difficulty Classifier}}

It is important to note that structural complexity alone does not fully determine the overall difficulty of an SQL query: some queries with numerous components may be logically straightforward and thus easier for models to generate correctly. From the perspective of LLMs, a more meaningful measure of difficulty is whether the model can consistently produce the correct SQL for a given NL question in the Text-to-SQL task. To this end, we introduce the \textit{execution difficulty} metric \( ed_i \). Specifically, we prompt the LLM to perform the Text-to-SQL task \( k \) times for the same input prompt $p_i$ and count the number of successful executions \( n \). The execution difficulty is then defined as:

\begin{equation}
    ed_i = 1 - \frac{n}{k}
\end{equation}

This metric reflects the practical capability of the current LLM in solving a specific Text-to-SQL problem. Unlike the component-based classifier, execution difficulty is model-dependent. More capable LLMs will achieve higher success rates on the same task, thereby exhibiting lower execution difficulty. Consequently, this metric provides a dynamic and performance-oriented perspective on task difficulty, better aligned with real-world generation behavior.

\subsection{Pipeline Workflow}

In the complete pipeline, the two classifier operators are incorporated as integral components. For each generated SQL query, the SQL Component Classifier and the SQL Execution Classifier are applied to assign difficulty labels to the corresponding data sample. Through this process, the pipeline produces data samples in the following format:

\begin{align*}
    \left\{ s_i^{\text{aug}},\, q_i,\, S,\, p_i,\, \text{cot}_i,\,  \text{ed}_i,\, \text{cd}_i \right\}_{i=1}^{N}
\end{align*}

\section{Details of Database Interaction Implementation}
\label{appendix:database_manager}

Within the framework of \textsc{Text2SQL-Flow}, a robust and efficient database interaction mechanism constitutes a core infrastructural component that underpins stable and scalable pipeline execution. Multiple critical stages of the pipeline, including SQL execution filtering and SQL augmentation, require frequent and reliable access to the underlying database system. These interactions can be broadly categorized into two types: (1) \textit{SQL Execution} and (2) \textit{Schema Metadata Extraction}.

\textit{SQL Execution} refers to submitting generated SQL queries to the database engine and collecting execution feedback, which is primarily used to validate the executability of candidate queries. In practice, mainstream database systems such as SQLite, MySQL, and PostgreSQL expose heterogeneous driver interfaces and connection protocols. Directly relying on database-specific native APIs complicates cross-platform support and requires maintaining separate adaptation logic for each backend, leading to code redundancy and severely limiting system extensibility.

\textit{Schema Metadata Extraction} aims to retrieve structured information about the database schema, including \texttt{CREATE TABLE} definitions, column attributes (e.g., data types, nullability, default values), primary and foreign key constraints, index metadata, and representative \texttt{INSERT} statements used for SQL augmentation. Such metadata is essential for understanding database semantics, generating syntactically and semantically valid SQL queries, and enabling context-aware data augmentation strategies. Although most relational databases expose schema information via mechanisms such as \texttt{INFORMATION\_SCHEMA} or system catalog tables, their query syntax, naming conventions, and output formats vary substantially across systems. Moreover, for a given database instance, schema information typically remains stable over short time horizons. Repeatedly querying identical metadata therefore introduces unnecessary I/O overhead and latency, which can significantly degrade pipeline throughput at scale.

To address these challenges, we design and implement a \texttt{Database Manager} module as part of the data augmentation framework. This module abstracts low-level database interaction details and exposes a unified, efficient, and extensible programming interface to upper-layer components. Specifically, the Database Manager supports \texttt{batch\_sql\_execution} and \texttt{batch\_compare\_sql}, enabling higher throughput under high-concurrency workloads. It also encapsulates schema metadata retrieval and provides high-level interfaces such as \texttt{get\_ddl} (for Data Definition Language extraction) and \texttt{get\_insert\_statement} (for generating representative insertion statements), thereby decoupling upper-layer logic from database-specific handling.

To ensure robust cross-database compatibility, we introduce an abstract base class, \texttt{DatabaseConnector}, which defines a standardized interaction interface. This interface includes methods such as \texttt{connect\_db} (for connection establishment and initialization), \texttt{execute\_sql} (for executing arbitrary SQL statements), and \texttt{get\_schema} (for extracting complete schema metadata). Concrete connectors for specific database systems (e.g., \texttt{SQLiteDatabaseConnector}) inherit from this base class and implement database-specific driver calls and error handling. At runtime, the Database Manager dynamically instantiates the appropriate connector based on configuration, while its core logic remains agnostic to the underlying database type. This design achieves the architectural principle of ``design once, adapt to many databases.''

To further improve throughput in large-scale SQL generation and validation scenarios, the Database Manager caches each database’s structured schema representation in memory. Although individual metadata queries are relatively fast, the pipeline repeatedly accesses the same schema when generating and validating multiple SQL candidates per database, making the cumulative overhead non-trivial.
Upon the first access to a database, its schema (including tables, columns, keys, and sampled rows) is extracted, serialized, and stored in an in-memory cache. Subsequent accesses are served directly from the cache, avoiding repeated schema reconstruction and redundant metadata queries.

In summary, the Database Manager decouples the pipeline from underlying database implementations, optimizes execution efficiency, and enhances system modularity. Its extensible, plugin-style architecture lays a solid foundation for supporting additional database types in the future.

\section{Case Study of Few-shot Retrieval}

\begin{figure}[t] 
\centering
\includegraphics[width=1.0\columnwidth]{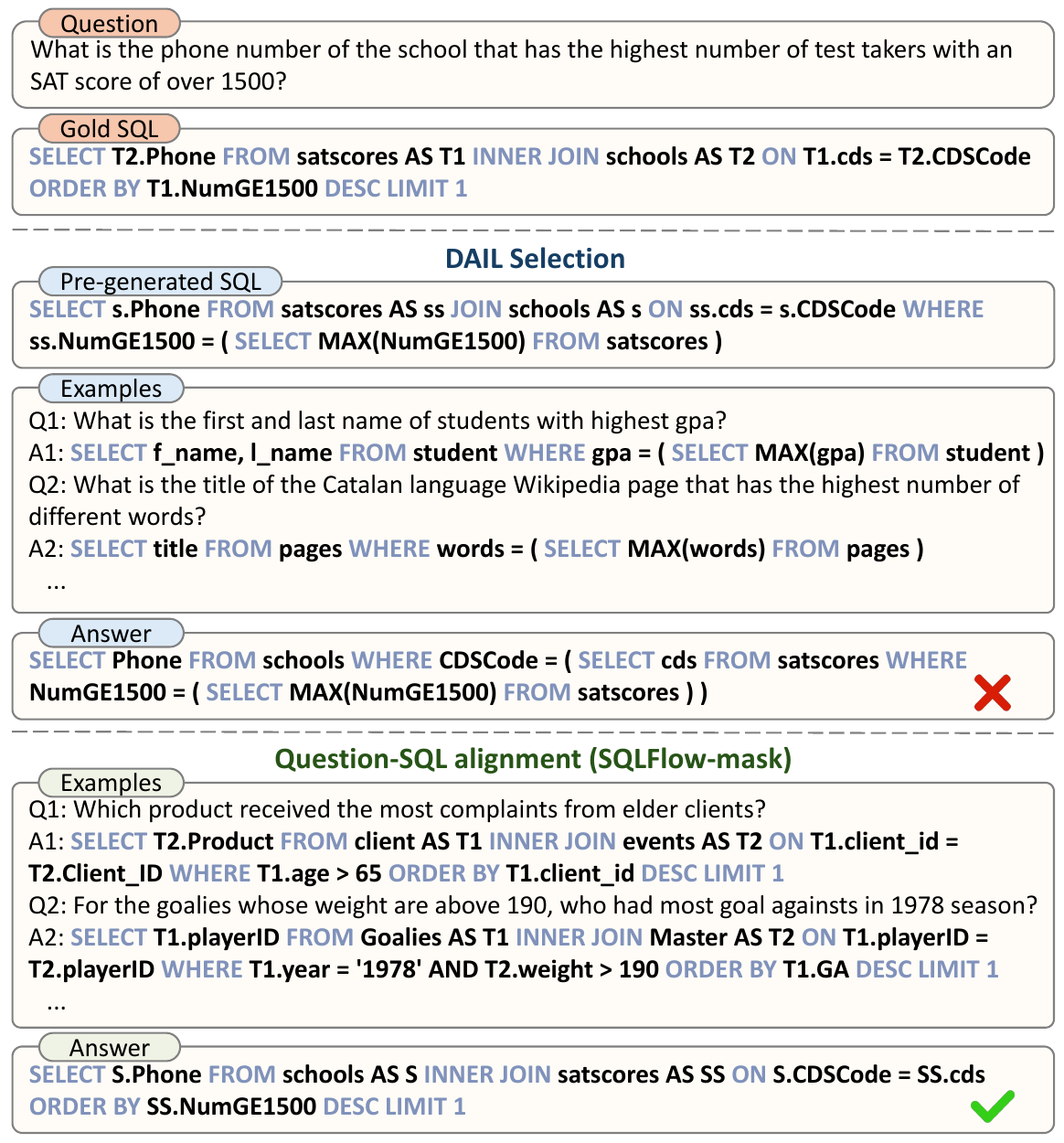}
\caption{Case study of few-shot retrieval.}
\label{fig:case_study}
\end{figure}

To illustrate how our method retrieves superior examples and why these examples effectively guide the model toward generating correct SQL, we present a case study from the BIRD-dev dataset, which is shown in Figure~\ref{fig:case_study}. DAIL Selection’s retrieval mechanism heavily relies on its pre-generated SQL. However, in this particular case, the pre-generated query is inaccurate both structurally and semantically. It fails to capture the key complex logic present in the Gold SQL, namely, returning only a single maximum-value result. Consequently, although the examples retrieved by DAIL Selection align structurally with this erroneous pre-generated query, they are entirely unrelated to the skeleton of the Gold SQL. These structurally flawed and irrelevant examples ultimately mislead the LLM, causing it to produce an incorrect answer. 
In contrast, our method directly aligns the question with the SQL skeleton, enabling it to retrieve examples highly similar to the Gold SQL skeleton using only the original question. As shown in the figure, the examples retrieved by our approach correctly include essential structural components such as \texttt{INNER JOIN}, \texttt{ORDER BY ... DESC}, and \texttt{LIMIT 1}. This demonstrates that our skeleton alignment mechanism captures not only syntactic similarity but also the core logical intent of the query. These high-quality examples provide the LLM with an accurate structural template, thereby guiding it to generate the correct SQL query.